\documentclass[12pt]{colt2014} 

\title[Lipschitz Bandits]{Lipschitz Bandits:\\ Regret Lower Bounds and Optimal Algorithms}
\usepackage{times,bbm}
\usepackage{algorithm,algorithmic,bbm}

\newtheorem{assumption}{Assumption}

\newcommand{\EE}{\mathbb{E}}
\newcommand{\NN}{\mathbb{N}}
\newcommand{\PP}{\mathbb{P}}
\newcommand{\RR}{\mathbb{R}}
\newcommand{\indic}{{\bf 1}}

\newcommand{\bp}{\noindent{\textbf{Proof.}}\ }
\newcommand{\ep}{\hfill $\Box$}

\newcommand{\al}[1]{ \begin{align} #1  \end{align}}
\newcommand{\eq}[1]{ \begin{equation} #1  \end{equation}}
\newcommand{\als}[1]{ \begin{align*} #1  \end{align*}}
\newcommand{\eqs}[1]{ \begin{equation*} #1  \end{equation*}}

\newcommand{\sk}{\nonumber\\}

\newcommand{\el}{\end{flushleft}}
\newcommand{\bl}{\begin{flushleft}}
 
\newcommand{\ceil}[1]{  \lceil #1 \rceil}

\newcommand{\separator}{
  \begin{center}
    \rule{\columnwidth}{0.3mm}
  \end{center}
}

\DeclareGraphicsExtensions{.pdf,.eps}

\graphicspath{{.//}}

\begin{document} 

\coltauthor{\Name{Stefan Magureanu} \Email{magur@kth.se}\\
 \addr KTH, The Royal Institute of Technology, EE School / ACL, Osquldasv. 10, Stockholm 100-44, Sweden  
 \AND
\Name{Richard Combes}\Email{richard.combes@supelec.fr}\\
\addr Supelec, Plateau de Moulon, 3 rue Joliot-Curie
91192 Gif-sur-Yvette Cedex, France
\AND
 \Name{Alexandre Proutiere} \Email{alepro@kth.se}\\
 \addr KTH, The Royal Institute of Technology, Stockholm, Sweden, and INRIA, Paris, France
 }

\maketitle

\begin{abstract}

\end{abstract}
We consider stochastic multi-armed bandit problems where the expected reward is a Lipschitz function of the arm, and where the set of arms is either discrete or continuous. For discrete Lipschitz bandits, we derive asymptotic problem specific lower bounds for the regret satisfied by any algorithm, and propose OSLB and CKL-UCB, two algorithms that efficiently exploit the Lipschitz structure of the problem. In fact, we prove  that OSLB is asymptotically optimal, as its asymptotic regret matches the lower bound. The regret analysis of our algorithms relies on a new concentration inequality for weighted sums of KL divergences between the empirical distributions of rewards and their true distributions. For continuous Lipschitz bandits, we propose to first discretize the action space, and then apply OSLB or CKL-UCB, algorithms that provably exploit the structure efficiently. This approach is shown, through numerical experiments, to significantly outperform existing algorithms that directly deal with the continuous set of arms. Finally the results and algorithms are extended to contextual bandits with similarities.

\section{Introduction}


In their seminal paper, \cite{lai1985} solve the classical stochastic Multi-Armed Bandit (MAB) problem. In this problem, the successive rewards of a given arm are i.i.d., and the expected rewards of the various arms are not related. They derive an asymptotic (when the time horizon grows large) lower bound of the regret satisfied by any algorithm, and present an algorithm whose regret matches this lower bound. This initial algorithm was quite involved, and many researchers have, since then,  tried to devise simpler and yet efficient algorithms. The most popular of these algorithms are UCB \cite{auer2002} and its extensions, e.g. KL-UCB \cite{garivier2011}, \cite{cappe2012} -- note that the KL-UCB algorithm was initially proposed and analysed in \cite{lai1987}, see (2.6). When the expected rewards of the various arms are not related as in  \cite{lai1985}, the regret of the best algorithm essentially scales as $O(K\log(T))$ where $K$ denotes the number of arms, and $T$ is the time horizon. When $K$ is very large or even infinite, MAB problems become more challenging. Fortunately, in such scenarios, the expected rewards often exhibit some structural properties that the decision maker can exploit to design efficient algorithms. Various structures have been investigated in the literature, e.g., Lipschitz \cite{agrawal95}, \cite{kleinberg2008}, \cite{bubeck08}, linear \cite{dani08}, and convex \cite{kalai05}. 

In this paper, we revisit bandit problems where the expected reward is a Lipschitz function of the arm. The set of arms is a subset of $[0,1]$ and we address both discrete Lipschitz bandits where this set is finite, and continuous Lipschitz bandits where this set is [0,1]. For discrete Lipschitz bandits, we derive problem specific regret lower bounds, and propose OSLB (Optimal Sampling for Lipschitz Bandits), an algorithm whose regret matches our lower bound. Most previous work on Lipschitz bandit problems address the case where the set of arms is [0,1], \cite{agrawal95}, \cite{kleinberg2008}, \cite{bubeck08}. For these problems, there is no known problem specific regret lower bound. In \cite{kleinberg2008}, a regret lower bound is derived for the {\it worst} Lipschitz structure. The challenge in the design of efficient algorithms  for continuous Lipschitz bandits stems from the facts that such algorithms should adaptively select a subset of arms to sample
from, and based on the observed samples, establish tight confidence
intervals and construct arm selection rules that optimally exploit the Lipschitz structure revealed by past observations. The algorithms proposed in \cite{agrawal95}, \cite{kleinberg2008}, \cite{bubeck08} adaptively define the set of arms to play, but used simplistic UCB indexes to sequentially select arms. In turn, these algorithms fail at exploiting the problem structure revealed by the past observed samples. For continuous bandits, we propose to first discretize the set of arms (as in \cite{kleinberg2008}), and then apply OSLB, an algorithm that optimally exploits past observations and hence  the problem specific structure. As it turns out, this approach outperforms algorithms directly dealing with continuous sets of arms.

\medskip  
\noindent  
{\bf Our contributions.} 

(a) For discrete Lipschitz bandit problems, we derive an asymptotic regret lower bound satisfied by any algorithm. This bound is problem specific in the sense that it depends in an explicit manner on the expected rewards of the various arms (this contrasts with existing lower bounds for continuous Lipschitz bandits).

(b) We propose OSLB (Optimal Sampling for Lipschitz Bandits), an algorithm whose regret matches our lower bound. We further present CKL-UCB (Combined KL-UCB), an algorithm that exhibits lower computational complexity than that of OSLB, and that is yet able to exploit the Lipschitz structure.

(c) We provide a finite time analysis of the regret achieved under OSLB and CKL-UCB. The analysis relies on a new concentration inequality for a weighted sum of KL divergences between the empirical distributions of rewards and their true distributions. We believe that this inequality can be instrumental for various bandit problems with structure.

(d) We evaluate our algorithms using numerical experiments for both discrete and continuous sets of arms. We compare their performance to that obtained using existing algorithms for continuous bandits. 

(e) We extend our results and algorithms to the case of contextual bandits with similarities as investigated in \cite{slivkins11}. 

\section{Models}

We consider a stochastic multi-armed bandit problem where the set of arms is a subset $\{x_1,\ldots,x_K\}$ of the interval $[0,1]$. Results can be easily extended to the case where the set of arms is a subset of a metric space as considered in \cite{kleinberg2008}. The set of arms is of finite cardinality, possibly large, and we assume without loss of generality that $x_1<x_2<\ldots<x_K$. Problems with continuous sets of arms are discussed in Section \ref{sec:num}. Time proceeds in rounds indexed by $n=1,2,\ldots$. At each round, the decision maker selects an arm, and observes the corresponding random reward. Arm $x_k$ is referred to as arm $k$ for simplicity. For any $k$, the reward of arm $k$ in round $n$ is denoted by $X_k(n)$, and the sequence of rewards $(X_k(n))_{n\ge 1}$ is i.i.d. with Bernoulli distribution of mean $\theta_k$ (the results can be generalized to distributions belonging to a certain parametrized family of distributions, but to simplify the presentation, we restrict our attention to Bernoulli rewards). The vector $\theta=(\theta_1,\ldots,\theta_K)$ represents the expected rewards of the various arms. Let ${\cal K}=\{1,\ldots,K\}$. We denote by $\theta^\star=\max_{k\in {\cal K}} \theta_k$ the expected reward of the best arm. A sequential selection algorithm $\pi$ selects in round $n$ an arm $k^\pi(n)\in {\cal K}$ that depends on the past observations. In other words, for any $n\ge 1$, if ${\cal F}_n^\pi$ denotes the $\sigma$-algebra generated by $(k^\pi(t),X_{k^\pi(t)}(t))_{1\le t\le n}$, then $k^\pi(n+1)$ is ${\cal F}_n^\pi$-measurable. Let $\Pi$ denote the set of all possible sequential selection algorithms. 

We assume that the expected reward is a Lipschitz function of the arm, and this structure is known to the decision maker. More precisely, there exists a positive constant $L$ such that for all pairs of arms $(k,k')\in {\cal K}$,
\begin{equation}\label{eq:lip}
|\theta_k -\theta_{k'}|\le L\times |x_k-x_{k'}|.
\end{equation}
We assume that $L$ is also known. We denote by $\Theta_L$ the set of vectors in $[0,1]^K$ satisfying (\ref{eq:lip}). The objective is to devise an algorithm $\pi\in \Pi$ that maximizes the average cumulative reward up to a certain round $T$ referred to as the time horizon ($T$ is typically large). Such an algorithm should optimally exploit the Lipschitz structure of the problem. As always in bandit optimization, it is convenient to quantify the performance of an algorithm $\pi\in \Pi$ through its expected regret (or regret for short) defined by:
$$
R^\pi(T) = T\theta^\star - \mathbb{E}[\sum_{n=1}^T X_{k^\pi(n)}(n)].
$$
  
\section{Regret Lower Bound}

\begin{figure}
\begin{center}
\includegraphics[width=0.6\columnwidth]{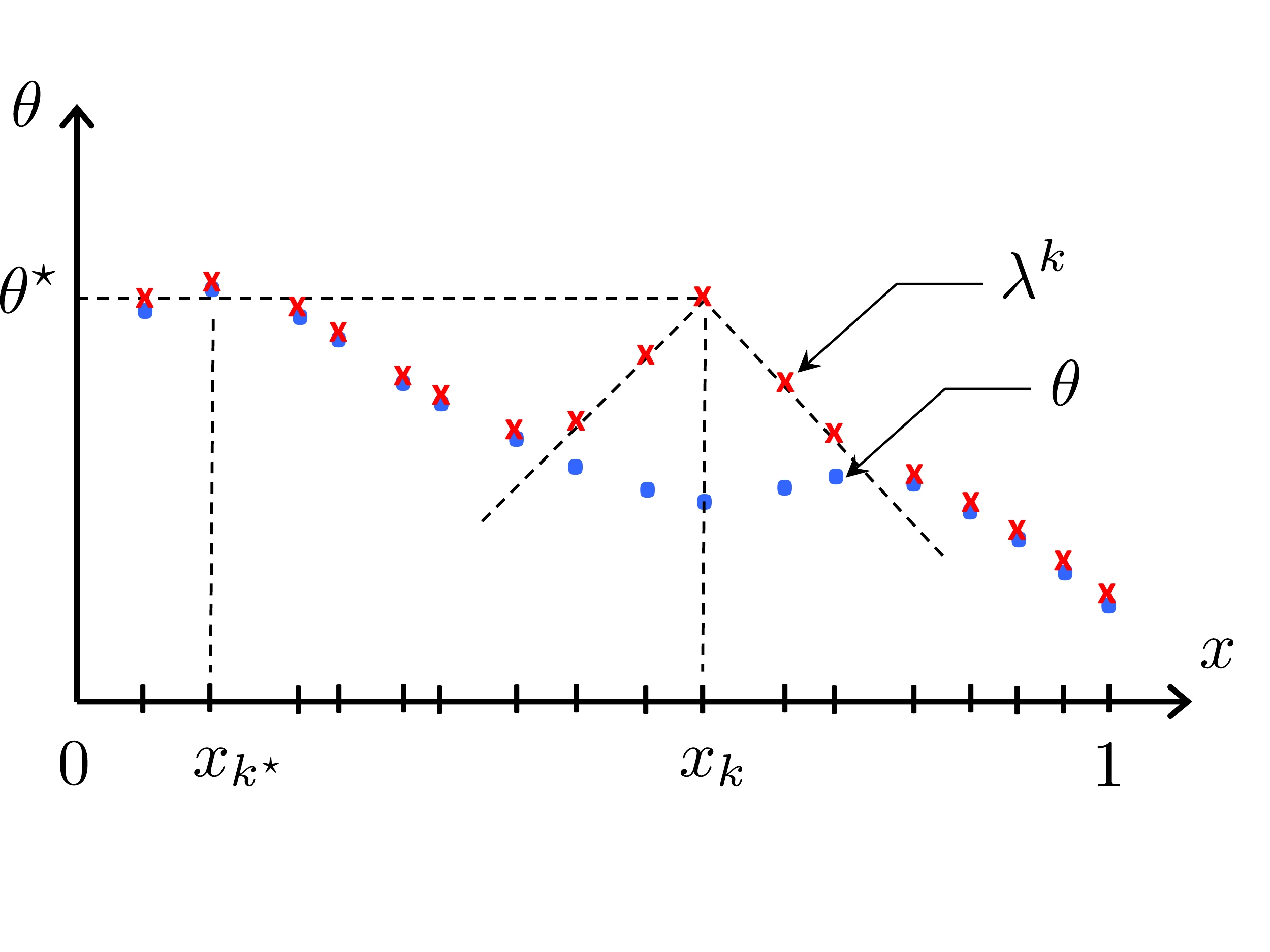}
\caption{A Lipschitz bandit with 17 arms, and an example of sub-optimal expected reward vector $\lambda^k$ involved in the regret lower bound.}
\end{center}
\label{fig:lambda1}
\end{figure}

In this section, we derive an asymptotic (when $T$ grows large) regret lower bound satisfied by any algorithm $\pi\in\Pi$. We denote by $I(x,y)=x\log({x\over y})+(1-x)\log({1-x\over 1-y})$ the KL divergence between two Bernoulli distributions with respective means $x$ and $y$. Fix the average reward vector $\theta=(\theta_1,\ldots,\theta_K)$. Let ${\cal K}^-=\{k\in {\cal K}:\theta_k<\theta^\star\}$ be the set of sub-optimal arms. For any $k\in {\cal K}^-$, we define $\lambda^k=(\lambda_1,\ldots,\lambda_K)$ as: $\forall i\in {\cal K}, \quad \lambda^k_i = \max\{ \theta_i, \theta^\star - L| x_k-x_i |\}$. The expected reward vector $\lambda^k$ is illustrated in Figure \ref{fig:lambda1}, and may be interpreted as the \textit{most confusing} reward vector among vectors in $\Theta_L$ such that arm $k$ (which is sub-optimal under $\theta$) is optimal under $\lambda^k$. This interpretation will be made clear in the proof of the following theorem. Without loss of generality, we restrict our attention to so-called \textit{uniformly good} algorithms, as defined in \cite{lai1985}. $\pi\in \Pi$ is uniformly good if for all $\theta\in \Theta_L$, $R^\pi(T)=o(T^a)$ for all $a>0$. Uniformly good algorithms exist -- for example, the UCB algorithm is uniformly good.

\begin{theorem}\label{th:low}
Let $\pi\in \Pi$ be a uniformly good algorithm. For any $\theta\in \Theta_L$, we have:
\begin{equation}\label{eq:low1}
\lim\inf_{T\to\infty} {R^\pi(T)\over \log(T)} \ge C(\theta),
\end{equation}
where $C(\theta)$ is the minimal value of the following optimization problem:
\begin{align}
& \min_{c_k\ge 0, \forall k\in {\cal K}^-} \sum_{k\in {\cal K}^-} c_k\times (\theta^\star-\theta_k)\label{eq:opt1}\\
& \hbox{s.t. } \forall k\in {\cal K}^-, \ \sum_{i\in {\cal K}} c_i I(\theta_i,\lambda^k_i) \ge 1.\label{eq:constraint}
\end{align}
\end{theorem}

The regret lower bound is a consequence of results in optimal control of Markov chains, see \cite{graves1997}. All proofs are presented in appendix. As in classical bandits, the minimal regret scales logarithmically with the time horizon. Observe that the lower bound (\ref{eq:low1}) is smaller than the lower bound derived in \cite{lai1985} when the various average rewards $(\theta_k, k\in {\cal K})$ are not related (i.e., in absence of the Lipschitz structure). Hence (\ref{eq:low1}) quantifies the gain one may expect by designing algorithms optimally exploiting the structure of the problem. Note that for any $k\in {\cal K}^-$, the variable $c_k$ corresponding to a solution of (\ref{eq:opt1}) characterizes the number of times arm $k$ should be played under an optimal algorithm: arm $k$ should be roughly played $c_k \log(n)$ times up to round $n$. 

It should be also observed that our lower bound is problem specific (it depends on $\theta$), which contrasts with existing lower bounds for continuous Lipschitz bandits, see e.g. \cite{kleinberg2008}. The latter are typically derived by selecting the problems that yield maximum regret. However, our lower bound is only valid for bandits with a finite set of arms, and cannot easily be generalized to problems with continuous sets of arms. 

\section{Algorithms}

In this section, we present two algorithms for discrete Lipschitz bandit problems. The first of these algorithms, referred to as OSLB (Optimal Sampling for Lipschitz Bandits), has a regret that matches the lower bound derived in Theorem \ref{th:low}, i.e., it is asymptotically optimal. OSLB requires that in each round, one solves an LP similar to (\ref{eq:opt1}). The second algorithm, CKL-UCB (Combined KL-UCB) is much simpler to implement, but has weaker theoretical performance guarantees, although it provably exploits the Lipschitz structure.

\subsection{The OSLB Algorithm}

To formally describe OSLB, we introduce the following notations. For any $n\ge 1$, let $k(n)$ be the arm selected under OSLB in round $n$. $t_k(n)$ denotes the number of times arm $k$ has been selected up to round $n-1$. By convention, $t_k(1)=0$. The empirical reward of arm $k$ at the end of round $(n-1)$ is $\hat{\theta}_k(n)={1\over t_k(n)}\sum_{t=1}^{n-1} \indic \{ k(t)=k  \} X_k(t)$, if $t_k(n) > 0$ and $\hat \theta_k(n) = 0$ otherwise. We denote by $L(n) = \arg \max_{k \in {\cal K}} \hat \theta_k(n)$ the arm with the highest empirical reward (ties are broken arbitrarily) at the end of round $n-1$. Arm $L(n)$ is referred to as the \textit{leader} for round $n$. We also define $\hat{\theta}^\star(n) = \hat\theta_{L(n)}(n)$ as the empirical reward of the leader at the end of round $n-1$. Let $f(n)=\log(n)+(3K+1)\log\log(n)$. Further define, for all $q\ge 0$ and $k$, the Lipschitz vector $\lambda^{q,k}$ such that for any $k'$, $\lambda_{k'}^{q , k} = q - L |x_k-x_{k'}|$. The sequential decisions made under OSLB are based on the {\it indexes} of the various arms. The index $b_k(n)$ of arm $k$ for round $n$ is defined by: 
$$
b_k(n)= \sup \{ q \in [\hat\theta_k(n),1]: \sk	
\sum_{k'=1}^{K} t_{k'}(n) I^{+}(\hat\theta_{k'}(n) ,  \lambda_{k'}^{q , k}) \leq f(n) \}.
$$
Note that the index $b_k(n)$ is always well defined, even for small values of $n$, e.g. $n=1$ (we have for all $x>0$, $I^+(0,x)=-\log(1-x)$).  
For any $\theta\in \Theta_L$, let $C(\theta)$ denote the minimal value of the optimization problem (\ref{eq:opt1}), and let $(c_k(\theta), k\in {\cal K}^-)$ be the values of the variables $(c_k, k\in {\cal K}^-)$ in (\ref{eq:opt1}) yielding $C(\theta)$. For simplicity, we define $\hat{C}(n) = C(\hat\theta(n))$, and $\hat{c}_k(n)=c_k(\hat\theta(n))$ for any $k\in {\cal K}^-(n)$ where ${\cal K}^-(n)=\{k: \hat\theta_k(n) < \hat\theta^\star(n)\}$. The design of OSLB stems from the observation that an optimal algorithm should satisfy $\lim_{n\to\infty}t_k(n)/(c_k(\theta)\log(n))=1$, almost surely, for all $k\in {\cal K}^-$. Hence we should force the exploration of arm $k\in {\cal K}^-(n)$ in round $n$ if $t_k(n)< \hat{c}_k(n)\log(n)$. We define the arm $\overline{k}(n)$ to explore as $\overline{k}(n) = \arg \min_{k\in K_e(n)} t_k(n)$ where $K_e(n)=\{k\in {\cal K}^-(n): t_k(n) \leq \hat c_k(n)\log(n) \}$. If $K_e(n)=\emptyset$, $\overline{k}(n) = -1$ (a dummy arm). Finally we define the least played arm as $\underline{k}(n) = \arg \min_{ k } t_k(n)$. In the definitions of $\overline{k}(n)$ and $\underline{k}(n)$, ties are broken arbitrarily. We are now ready to describe OSLB. Its pseudo-code is presented in Algorithm \ref{alg:oslb}.

\begin{algorithm}[t!]
   \caption{OSLB($\epsilon$)}
   \label{alg:oslb}
\begin{algorithmic}
\STATE For all $n\ge 1$, select arm $k(n)$ such that:
\STATE If $\hat{\theta}^\star(n) \geq \max_{k \neq L(n)} b_k(n)$, then  $k(n) = L(n)$;
\STATE Else If $t_{\underline{k}(n)}(n) < {\epsilon\over K} t_{\overline{k}(n)}(n)$, then $k(n) = \underline{k}(n)$;
\STATE  \hspace{7mm} Else  $k(n) = \overline{k}(n)$.
\end{algorithmic}
\end{algorithm}

Under OSLB, the leader is selected if its empirical average exceeds the index of other arms. If this is not the case, OSLB selects the least played arm $\underline{k}(n)$, if the latter has not been played enough, and arm $\overline{k}(n)$ otherwise. Note that the description of OSLB is valid in the sense that $\overline{k}(n) \neq -1$ if  $\hat{\theta}^\star(n) < \max_{k \neq L(n)} b_k(n)$. After each round, all variables are updated, and in particular $\hat{c}_k(n)$ for any $k\in {\cal K}^-(n)$, which means that at each round we solve an LP, similar to (\ref{eq:opt1}).

\subsection{The CKL-UCB Algorithm}

Next, we present the algorithm CKL-UCB (Combined KL - UCB). The sequential decisions made under CKL-UCB are based on the indexes $b_k(n)$, and CKL-UCB explores the apparently suboptimal arms by choosing the least played arms first. When the leader $L(n)$ has the largest index, it is played, and otherwise we play the arm in $\{ k: b_{k}(n) > b_{L(n)}(n)\}$, the set of arms which are possibly better than the leader, with the least number of current plays. Note that in practice, the forced $\log\log(n)$ exploration is unnecessary and only appears to aide in the regret analysis. 

The rationale behind CKL-UCB is that if we are given a set of suboptimal arms, by exploring them, we will first eliminate arms whose expected reward is low (these arms do not require many plays to be eliminated). Note that the arm chosen by CKL-UCB is directly computed from the indexes, without solving an LP, and hence CKL-UCB is computationally light. From a practical perspective, CKL-UCB should also be more robust than OSLB in the sense that it does not take decisions based on the solution of the LP calculated with empirical averages $\hat\theta(n)$. This could be problematic if the LP solution is very sensitive to errors in the estimate of $\theta$.

\begin{algorithm}[t!]
   \caption{CKL-UCB}
   \label{alg:RR-UCB}
\begin{algorithmic}
\STATE For all $n\ge 1$, select arm $k(n)$ such that:
\STATE If $\exists k$ such that $t_k(n) <\log\log(n)$, then k(n) = k (ties are broken arbitrarily);
\STATE Else if $b_{L(n)}(n) \geq \max\limits_{k \neq L(n)} b_k(n)$, then  $k(n) = L(n)$;
\item \hspace{7mm} Else
$k(n) = \arg\min\limits_{k}\{t_k(n): b_{k}(n) > b_{L(n)}(n)\}$ (ties are broken arbitrarily).

\end{algorithmic}
\end{algorithm}

\section{Regret Analysis}

In this section, we provide finite time upper bounds for the regret achieved under OSLB and CKL-UCB. 

\subsection{Concentration Inequalities}

To analyse the regret of algorithms for bandit optimization problems, one often has to leverage results related to the concentration-of-measure phenomenon. More precisely, here, in view of the definition of the indexes $b_k(n)$, we need to establish a concentration inequality for a weighted sum of KL divergences between the empirical distributions of rewards and their true distributions. We derive such an inequality. The latter extends to the multi-dimensional case the concentration inequality derived in~\cite{Garivier2013} for a single KL divergence. We believe that this inequality can be instrumental in the analysis of general structured bandit problems, as well as for statistical tests involving vectors whose components have distributions in a one-parameter exponential family (such as Bernoulli or Gaussian distributions). For simplicity, the inequality is stated for Bernoulli random variables only.

We use the following notations. For $k\in {\cal K}$, let $\{ X_k(n) \}_{n \in \NN}$ be a sequence of i.i.d. Bernoulli random variables with expectation $\theta_k$ and $X(n)= (X_k(n), k\in {\cal K})$. We represent the history up to round $n$ using the $\sigma$-algebra ${\cal F}_n = \sigma(X(1),\dots,X(n))$, and define the natural filtration ${\cal F} = \{ {\cal F}_n \}_{n \ge 1}$. We consider a generic sampling rule $B(n)=(B_k(n), k\in {\cal K})$ where $B_k(n) \in \{0,1\}$ for all $k\in {\cal K}$. The sampling rule is assumed to be predictable in the sense that $B(n) \in {\cal F}_{n-1}$.

We define the number of times that $k$ was sampled up to round $n-1$ by $t_k(n) = \sum_{t=1}^{n-1} B_k(t)$ and the sum $S_k(n) = \sum_{t=1}^{n-1} B_k(t) X_k(t)$. The empirical average for $k$ is $\hat\theta_k(n) = S_k(n)/t_k(n)$ if $t_k(n) > 0$ and $\hat\theta_k(n) = 0$ otherwise. Finally, we define the vectors $\hat\theta(n) = (\hat\theta_1(n), \dots , \hat\theta_K(n))$ and $t(n) = (t_1(n), \dots , t_K(n))$. When comparing vectors in $\RR^K$, we use the component-by-component order unless otherwise specified. 

\begin{theorem}\label{th:kl_concentr}
For all $\delta \geq (K+1)$ and $n \in \NN$ we have:
\begin{equation}\label{eq:con}
\PP\left[ \sum_{k=1}^K  t_k(n) I^{+}(\hat\theta_k(n) , \theta_k   ) \geq \delta \right] \leq e^{-\delta} \left( \frac{\ceil{\delta \log(n)} \delta}{K}\right)^K e^{K+1} .
\end{equation}
\end{theorem}

The proof of Theorem \ref{th:kl_concentr} involves tools that are classically used in the derivation of concentration inequalities, but also requires the use of stochastic ordering techniques, see e.g. \cite{MullerStoyan}.

\subsection{Finite time analysis of OSLB}

Next we provide a finite time analysis of the regret achieved under OSLB, under the following mild assumption. This assumption greatly simplifies the analysis.

\begin{assumption}\label{assum:uniqueLP}
The solution of the LP~\eqref{eq:opt1} is unique.
\end{assumption}

It should be observed that the set of parameters $\theta \in \Theta_L$ such that Assumption 1 is satisfied constitutes a dense subset of $\Theta_L$.

\begin{theorem}\label{th:oslbperf}
For all $\epsilon>0$, under Assumption \ref{assum:uniqueLP}, the regret achieved under $\pi=\text{OSLB}(\epsilon)$ satisfies: for all $\theta\in \Theta_L$, for all $\delta > 0$ and $T \ge 1$,  
\begin{equation}\label{eq:oslbperf}
R^{\pi}(T) \leq C^\delta(\theta) (1+ \epsilon) \log(T) + C_1 \log\log(T) + K^3 \epsilon^{-1} \delta^{-2}+3K\delta^{-2},
\end{equation}
where $C^\delta(\theta) \to C(\theta)$, as $\delta \to 0^+$, and $C_1>0$.
\end{theorem}

In view of the above theorem, when $\epsilon$ is small enough, OSLB($\epsilon$) approaches the fundamental performance limit derived in Theorem \ref{th:low}. More precisely, we have for all $\epsilon>0$ and $\delta>0$:
$$
\lim\sup_{T\to\infty} {R^\pi(T)\over \log(T)} \le C^\delta(\theta)(1+\epsilon).
$$
In particular, for any $\zeta>0$, one can find $\epsilon>0$ and $\delta>0$ such that $C^\delta(\theta)(1+\epsilon)\le (1+\zeta)C(\theta)$, and hence, under $\pi =$OSLB($\epsilon$),
$$
\lim\sup_{T\to\infty} {R^\pi(T)\over \log(T)} \le C(\theta)(1+\zeta).
$$

\subsection{Finite Time analysis of CKL-UCB}

In order to analyze the regret of  CKL-UCB, we define the following optimization problem.  Define the matrix of Kullback-Leibler divergence numbers $A = (a_{ik})_{i,k\in {\cal K}}$ with $a_{ik} =  I( \theta_i, \lambda^{k,\theta^\star}_i )$. Consider an arm $k \neq k^\star$, a subset of arms ${\cal N} \subset \{1,\dots,K\} \setminus \{ k,k^\star \}$, and $\alpha_0\geq 0$. We define $d_k(A,\alpha_0,{\cal N})$ the optimal value of the following linear program:
\als{   \min_{\alpha_1,\dots,\alpha_K} &  \sum_{k'  \in {\cal K}^- \setminus \{k\} } \alpha_{k'} a_{k'k} \sk
	\text{ s.t.\ }       &      \alpha_{k'} \geq \alpha_0,\;\;\;\forall k' \not\in {\cal N} , k' \neq k^\star \sk
	 										&		 \alpha_{k'} \geq 0,\;\;\; \forall k' \sk
	 									&			 \sum_{k'' \in {\cal K}^- \setminus \{k\} } \alpha_{k''} a_{k''k'} \geq  1 - \alpha_0 a_{kk'},\;\;\;\forall k' \in {\cal N}.
}
and $e_k(A,\alpha_0) = \min_{ {\cal N} }  d_k(A,\alpha_0,{\cal N})$ where the minimum is taken over all possible subsets of $\{1,\dots,K\} \setminus \{ k,k^\star \}$.

\begin{theorem}\label{th:rrucbperf}
Under CKL-UCB, for all $\theta \in \Theta_{L}$, all $T \geq 1$, all $0 < \delta < (\theta^\star - \max_{k \neq k^\star} \theta_k)/2$, and any suboptimal arm $k \in {\cal K}^{-}$, \\
(i) we have:
\eqs{
\EE[ t_k(T) ] \leq \frac{f(T)}{I(\theta_k+\delta, \theta^*-\delta)} + C_1 \log(\log(T)) +  2 \delta^{-2}.
}
with $C_1 \geq 0$ a constant.\\
(ii) Furthermore, for all $k \in {\cal K}^-$, we have that:
\eqs{
\lim \sup_{T \to \infty} \frac{\EE[t_k(T)]}{\log(T)} \leq \beta_k( \theta ).
}
where
\eqs{
	\beta_k(\theta ) = \inf \{ \alpha_0 \geq 0:  a_{k,k} \alpha_0 + e_k(A,\alpha_0)  > 1    \}.
}
(iii) Assume that there exists $k'$ such that $0 < a_{kk'} < a_{kk}$ and such that for all $k''$ we have that if $a_{k''k} = 0$ then $a_{k''k'}=0$ as well. Then $\beta_k(\theta) < 1/a_{kk} = 1/I(\theta_k,\theta^\star)$.
\end{theorem}

In the above theorem, statement (i) shows that CKL-UCB plays arm $k$ at most as much as KL-UCB, so that CKL-UCB outperforms KL-UCB for any value of the parameters $\theta$. Now statements (ii) and (iii) show that under certain assumptions, CKL-UCB plays arm $k$ strictly less than KL-UCB, so that CKL-UCB indeed exploits the Lipshitz structure of the problem. Note that the conditions in (iii) holds for triangular reward functions, and other unimodal functions, and hence in these cases, CKL-UCB strictly outperforms KL-UCB. The regret analysis of CKL-UCB presented above is preliminary, and we believe that its performance guarantees can be further improved.

\section{Contextual Bandit with Similarities}

The algorithms and results presented above can be extended to the case of contextual bandit problems with similarities as studied in \cite{slivkins11}. In such problems, in each round, the decision maker observes a context, and then decides which arm to select. The expected reward of the various arms depends on the context, and is assumed to be Lipschitz in the arm and context. We assume that contexts arrive according to an i.i.d. process whose distribution is not known to the decision maker. This contrasts with most of the work in contextual bandits, where the context process is adversarial.

\subsection{Model}
Let $\{y_1,\ldots,y_J\}$ denote the set of possible contexts, assumed to be a subset of $[0,1]$. We assume that $y_1<\ldots<y_J$. For simplicity, context $y_j$ is referred to as context $j$. For each context $j\in {\cal J}=\{1,\ldots,J\}$, the expected rewards of the various arms are represented by a vector $\theta(j)=(\theta_k(j),k\in {\cal K})$ ($\theta_k(j)$ is the expected reward of arm $k$ when the context is $j$). We consider a general scenario where the reward is a Lipschitz function in both the arm and the context. There exists $L$ (known to the decision maker) such that for all $(i,k),(j,l)\in {\cal J}\times {\cal K}$,
\begin{equation}\label{eq:dd}
| \theta_k(i)-\theta_l(j)|\le L\times {\cal D}((i,k),(j,l)),
\end{equation}
where ${\cal D}$ refers to some {\it metric} over ${\cal J}\times {\cal K}$. The choice of this metric is free, and allows us to consider different scenarios. For example, we may assume that the Lipschitz structure is stronger in terms of arms than in terms of contexts. In this case, we may choose, for some $\beta>1$, \\
${\cal D}((i,k),(j,l))=\sqrt{(\beta (y_i-y_j)^2 + (x_k-x_l)^2)}$ . The set of $\theta=(\theta_k(j), k\in {\cal K},j\in {\cal J})$ satisfying (\ref{eq:dd}) is denoted by $\Theta_{L,2}$.

The context process is i.i.d.. The distribution of the observed context $j(n)$ in round $n$ is $\psi$, i.e., $\psi(j)=\mathbb{P}[j(n)=j]$. Without loss of generality, we assume that for any $j\in {\cal J}$, $\psi(j)>0$. $\psi$ is unknown to the decision maker. Let $X_{j,k}(n)$ denote the reward of arm $k$ obtained in round $n$ when the context is $j$. For contextual bandits, we define the regret of algorithm $\pi$ as follows: 
\begin{equation}
R^\pi(T) = T \sum_{j \in \cal{J}} \psi(j)\theta^\star(j) - \sum_{n=1}^T \mathbb{E}[X_{j(n),k^\pi(n)}(n)].
\end{equation}
where $\theta^\star(j)$ denotes the reward of the best arm under context $j$, and as earlier $k^\pi(n)$ denotes the arm selected under $\pi$ in round $n$.

\subsection{Regret Lower Bound}

To state the regret lower bound, we introduce for any context $j\in {\cal J}$, ${\cal K}^-(j)=\{k\in {\cal K}:\theta_k(j)<\theta^\star(j)\}$ the set of suboptimal arms for context $j$. We also introduce for any context $j\in {\cal J}$, and any $k\in {\cal K}$, the vector $(\lambda_l^{j,k}(i), l\in {\cal K},i\in {\cal J})$ such that
$$
\lambda_l^{j,k}(i)=\max \{\theta_l(i), \theta^\star(j)- L{\cal D}((j,k),(i,l))\}.
$$

\begin{theorem}
Let $\pi$ be a uniformly good algorithm. Then, for any $\theta\in \Theta_{L,2}$:
\begin{equation}
\lim\inf\limits_{T\to\infty} \frac{R^{\pi}(T)}{\log(T)}\geq C'(\theta)
\end{equation}
where $C'(\theta)$ is the minimal value of the following optimization problem:
\begin{align}
& \min\limits_{c_{j,k}\geq 0,\forall j,\forall k} \quad\sum_{j\in{\cal J}}\sum_{k\in{\cal K^-}(j)} c_{j,k}\times (\theta^*(j) -  \theta_k(j))\\
&\hbox{s.t. } \forall j, \forall k\in {\cal K}^-(j), \quad \sum_{i\in{\cal J}}\sum_{l\in{\cal K}} c_{i,l}I(\theta_l(i), \lambda^{j,k}_l(i)) \geq 1.\label{const2}
\end{align}
\end{theorem}

Observe that our regret lower bound is problem specific, and again the values of the $c_{j,k}$'s solving the above optimization problem can be interpreted as follows: an asymptotically optimal algorithm plays arm $k$ when the context is $j$ a number of times that scales as $c_{j,k}\log(T)$ as $T$ grows large. Also note that the regret lower bound does not depend on the distribution $\psi$ of the contexts.


\subsection{Algorithms}

The algorithms proposed for Lipschitz bandits can be naturally extended to the case of contextual bandits with similarities. For conciseness, we just present CCKL-UCB (Contextual Combined KL - UCB), the extension of CKL-UCB. Its regret analysis can be conducted as that of CKL-UCB with minor modifications.

To describe CCKL-UCB, we introduce the following notations. Let $\hat\theta_k(j,n)$ denote  the empirical average reward of arm $k$ for context $j$ up to round $n-1$. $t_k(j,n)$ is the number of times context $j$ is presented and arm $k$ is chosen up to round $n-1$. We define the index $b_k^c(j,n)$ of arm $k$ for round $n$, when the context $j$ is observed as:
\eqs{
b_k^c(n,j) = \sup \{q\in [\hat\theta_k(j,n),1] : \sum\limits_{i\in {\cal J}} \sum\limits_{l\in{\cal K}} t_l(i,n) I^+(\hat\theta_l(i,n), \lambda^{q,k,j}_l(i,n)) \leq f(n)\},
}
where $\lambda^{q,k,j}_l(i,n)=q-L{\cal D}((j,k),(i,l))$. As for Lipschitz bandits,  the indexes are built so as to match the constraints (\ref{const2}) of the optimisation problem leading to the regret lower bound. The leader for round $n$ and context $j$ is defined $L(n,j)= \arg\max\limits_k \hat\theta_k(j,n)$ (ties are broken arbitrarily). In round $n$, CCKL-UCB plays the leader $L(n,j(n))$ for the current context if it has the highest index, and otherwise selects the least played arm which has an index higher than the leader $L(n,j(n))$.
\begin{algorithm}[t!]
   \caption{CCKL-UCB}
   \label{alg:CRR-UCB}
\begin{algorithmic}
\STATE For all $n\ge 1$, observe context $j=j(n)$, and select arm $k(n)$ such that:
\STATE If $\exists k$ such that $t_k(j,n) <\log\log(n)$, then k(n) = k (ties are broken arbitrarily);
\STATE Else if $L(n,j) = \arg\max\limits_k b^c_k(n,j)$, then  $k(n) = L(n,j)$;
\item \hspace{7mm} Else $k(n) = \arg\min\limits_{k} \{ t_k(j,n) : b^c_k(j,n) > b^c_{L(n)}(j,n)\}$ (ties are broken arbitrarily).
\end{algorithmic}
\end{algorithm}

\section{Numerical Experiments}\label{sec:num}

In this section, we present numerical experiments illustrating the performance of our algorithms compared to other existing algorithms.

\subsection{Discrete Lipschitz Bandits}

\begin{figure}[h]
\centering
  \includegraphics[width=0.45\linewidth]{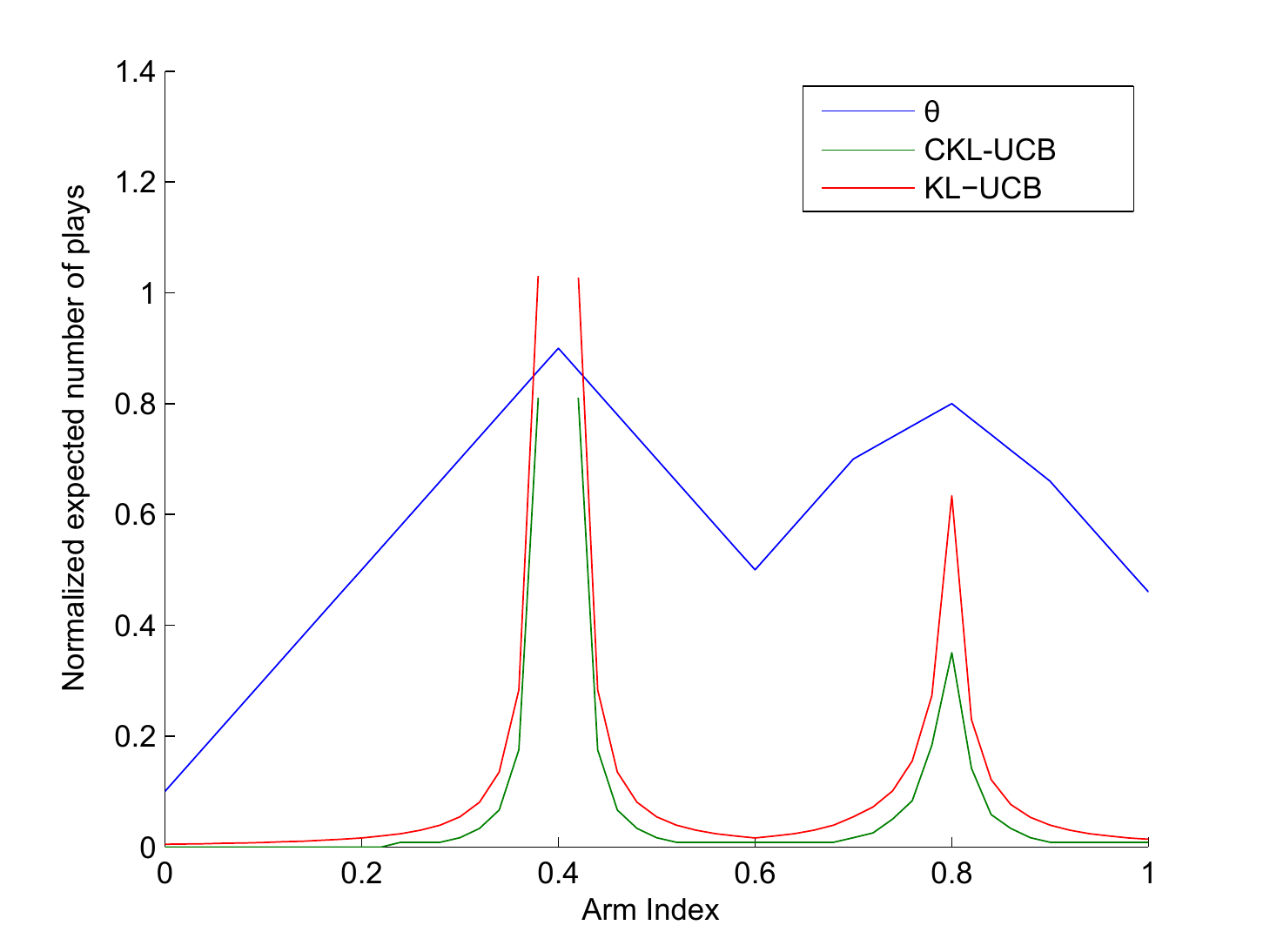}
    \includegraphics[width=0.45\linewidth]{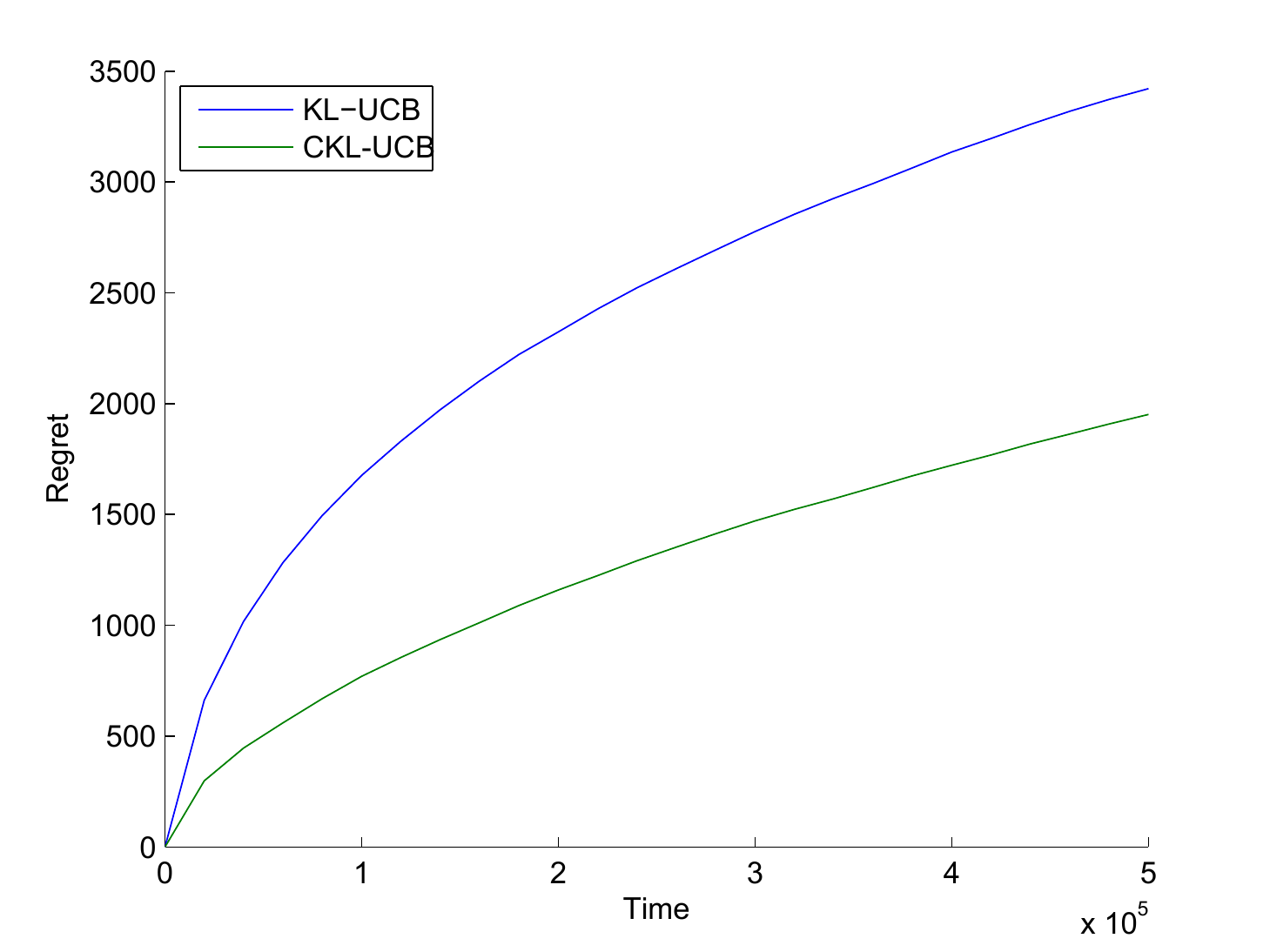}
  \caption{(Left) The expected rewards and the scaled amount of times suboptimal arms are played under KL-UCB and CKL-UCB as a function of the arm. (Right) Regret under KL-UCB and CKL-UCB as a function of time.}
  \label{fig:discrete2}
\end{figure}

We first consider discrete bandit problems with $46$ arms, and with time horizons less than $T=5.10^{5}$ rounds. The regret is averaged over $150$ runs. In Figure \ref{fig:discrete2}, we compare the performance of KL-UCB and CKL-UCB. For improved numerical performance, in the case of both algorithms we ignore the $\log\log(n)$ terms in the indexes (i.e.$f(n) = \log(n)$). On the left, we plot the expected reward as a function of the arm, as well as the (scaled) amount of times $\EE[t_k(n)]/\log(n)$ sub-optimal arm $k$ is played under both algorithms, as function of time. Under KL-UCB, the amount of times for arm $k$ approaches $1/I(\theta_k,\theta^\star)$, whereas under CKL-UCB, $\EE[t_k(n)]$ satisfy the upper bounds derived in Theorem \ref{th:rrucbperf}. CKL-UCB explores suboptimal arms less often than KL-UCB, as it is designed to exploit the Lipschitz structure. On the right, we plot the expected regret as a function of time under both algorithms. The regret under CKL-UCB is always smaller than that under KL-UCB (the regret under KL-UCB is typically twice as large as that under CKL-UCB in this example). This illustrates the significant gains that one may achieve by efficiently exploiting the structure of the problem.

\subsection{Continuous Lipschitz Bandits}
\begin{figure}[h]
\centering
\includegraphics[width=0.45\linewidth]{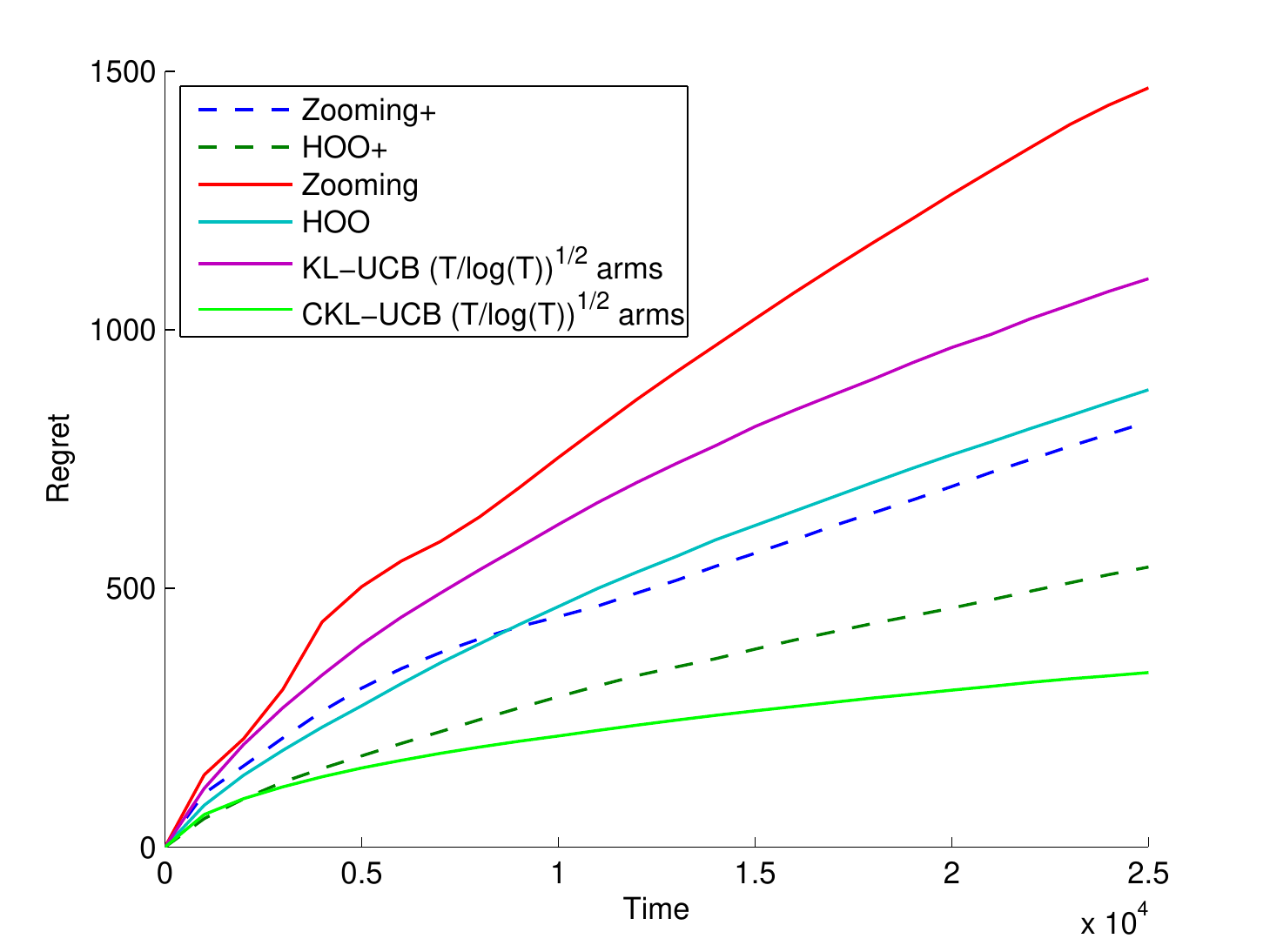}
\includegraphics[width=0.45\linewidth]{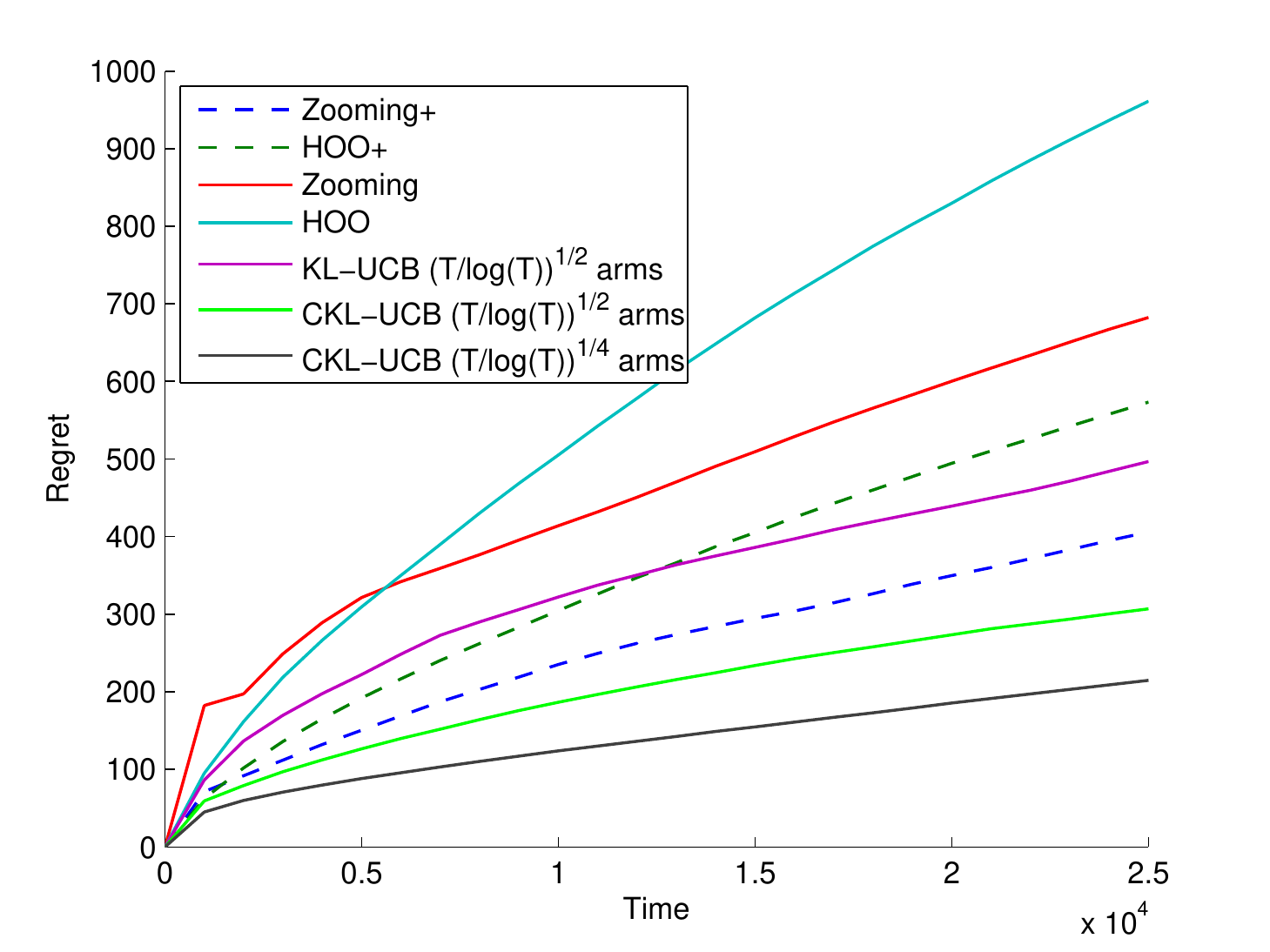}
\caption{Expected regret of different algorithms as function of time for a triangular reward function (left) and a quadratic reward function (right).}
\label{fig:figureContinuous}
\end{figure}

We now turn our attention to continuous Lipschitz bandits where the set of arms is [0,1]. We consider two reward functions that behave differently around their maximum: (1) $\theta(x) = 0.8 - 0.5 |0.5 - x|$ (triangle) and (2) $\theta(x) = \max(0.1, 0.9- 3.2* (0.7 - x)^2)$ (quadratic function). To adapt KL-UCB and CKL-UCB to this continuous setting, we use a uniform discretization of the set of arms, with $\delta^{-1}=\lceil \sqrt{T/\log(T)}\rceil$ arms. This discretization is known to be order-optimal for functions which are regular around their maximum~\cite{Kleinberg2004}. In order not to give a positive bias to KL-UCB and CKL-UCB, we make sure that the maximum of the reward functions is not achieved in one of the arms in the discretization: the maximum is placed at a distance of at least $\delta/4$ from any arm in the discretization. We compare the performance of KL-UCB and CKL-UCB to that of the algorithm HOO introduced in \cite{bubeck08}, and the Zooming algorithm proposed in \cite{kleinberg2008}. The two latter algorithms have performance guarantees (they are order-optimal). We also compare KL-UCB and CKL-UCB to HOO+ and Zooming+, two improved versions of HOO and Zooming, respectively. In these tuned versions, the {\it confidence radius} (see \cite{bubeck08} and \cite{kleinberg2008} for details) is set equal to $\sqrt{\log(n)/(2*t_k(n))}$ in round $n$. HOO+ and Zooming+ exhibit better performance than their initial versions, but their regrets have not been analytically studied. In the experiments, we limit the time horizon to $T=25 000$ rounds, and the expected regret is calculated by averaging over $100$ independent runs.

Figure \ref{fig:figureContinuous} presents the expected regret of the various algorithms for the triangular reward function (left) and for the quadratic reward function (right). First note that surprisingly, KL-UCB, an algorithm that does not leverage the Lipschitz structure, outperforms some of the algorithms designed to exploit the structure. Observe that CKL-UCB clearly outperforms KL-UCB and all other algorithms in both problem instances. For quadratic reward functions, it is known that the optimal discretization of the set of arms should roughly have $(\log(T)/T)^{1/4}$ arms, \cite{Combes2014}. We also plot the regret achieved under CKL-UCB using this optimized discretization, and we observe that this indeed further reduces the regret.   

It is worth noting that in the case of CKL-UCB most of the regret is caused by not discretizing enough around the top arm. In contrast, in the case of Zooming and HOO, most of the regret is caused by loose confidence bounds. Therefore, in future work we will explore the possibility of combining the adaptive discretization scheme of Zooming and HOO with efficient confidence bounds as used by CKL-UCB.

\section{Conclusion}
We consider stochastic multi-armed bandits (discrete or continuous) where the expected reward is a Lipschitz function of the arm. For discrete Lipschitz bandits, we derive asymptotic lower bounds for the regret achieved under any algorithm. We propose OSLB and CKL-UCB, two algorithms that exploit the Lipschitz structure efficiently.  OSLB is asymptotically optimal and CKL-UCB is a computationally light algorithm  which efficiently exploits the Lipschitz structure. The regret analysis is based on a new concentration inequality for sums of KL divergences which can be instrumental for bandit problems with correlated arms. For continuous Lipschitz bandits, we adapt OSLB and CKL-UCB by using a simple discretization. For both discrete and continuous bandits, initial numerical experiments show that our approach significantly outperforms the state-of-the-art algorithms. Finally the results and algorithms are extended to contextual bandits with similarities.

\clearpage
\newpage

\bibliography{RA}

\begin{thebibliography}{18}
\providecommand{\natexlab}[1]{#1}
\providecommand{\url}[1]{\texttt{#1}}
\expandafter\ifx\csname urlstyle\endcsname\relax
  \providecommand{\doi}[1]{doi: #1}\else
  \providecommand{\doi}{doi: \begingroup \urlstyle{rm}\Url}\fi

\bibitem[Agrawal(1995)]{agrawal95}
R.~Agrawal.
\newblock The continuum-armed bandit problem.
\newblock \emph{SIAM J. Control and Optimization}, 33\penalty0 (6):\penalty0
  1926--1951, November 1995.

\bibitem[Auer et~al.(2002)Auer, Cesa-Bianchi, and Fischer]{auer2002}
P.~Auer, N.~Cesa-Bianchi, and P.~Fischer.
\newblock Finite time analysis of the multiarmed bandit problem.
\newblock \emph{Machine Learning}, 47\penalty0 (2-3):\penalty0 235--256, 2002.

\bibitem[Bubeck et~al.(2008)Bubeck, Munos, Stoltz, and Szepesv\'ari]{bubeck08}
S.~Bubeck, R.~Munos, G.~Stoltz, and C~Szepesv\'ari.
\newblock Online optimization in x-armed bandits.
\newblock In \emph{Advances in Neural Information Processing Systems 22}, 2008.

\bibitem[Capp\'e et~al.(2013)Capp\'e, Garivier, Maillard, Munos, and
  Stoltz]{cappe2012}
O.~Capp\'e, A.~Garivier, O.~Maillard, R.~Munos, and G.~Stoltz.
\newblock Kullback-leibler upper confidence bounds for optimal sequential
  allocation.
\newblock \emph{Annals of Statistics}, 41\penalty0 (3):\penalty0 516--541, June
  2013.

\bibitem[Combes and Proutiere(2014{\natexlab{a}})]{Combes2014}
R.~Combes and A.~Proutiere.
\newblock Unimodal bandits: Regret lower bounds and optimal algorithms.
\newblock In \emph{Proc. of ICML}, 2014{\natexlab{a}}.

\bibitem[Combes and Proutiere(2014{\natexlab{b}})]{techreport}
R.~Combes and A.~Proutiere.
\newblock Unimodal bandits: Regret lower bounds and optimal algorithms.
\newblock Technical Report, people.kth.se/\~{}alepro/pdf/tr-icml2014.pdf,
  2014{\natexlab{b}}.

\bibitem[Dani et~al.(2008)Dani, Hayes, and Kakade]{dani08}
V.~Dani, T.~P. Hayes, and S.~M. Kakade.
\newblock Stochastic linear optimization under bandit feedback.
\newblock In \emph{Proc. of Conference On Learning Theory (COLT)}, pages
  355--366, 2008.

\bibitem[Flaxman et~al.(2005)Flaxman, Kalai, and McMahan]{kalai05}
A.~Flaxman, A.~T. Kalai, and H.~B. McMahan.
\newblock Online convex optimization in the bandit setting: gradient descent
  without a gradient.
\newblock In \emph{Proc. of {ACM/SIAM} symposium on {D}iscrete {A}lgorithms
  (SODA)}, pages 385--394, 2005.

\bibitem[Garivier(2013)]{Garivier2013}
A.~Garivier.
\newblock Informational confidence bounds for self-normalized averages and
  applications.
\newblock In \emph{Information Theory Workshop}, 2013.

\bibitem[Garivier and Capp\'e(2011)]{garivier2011}
A.~Garivier and O.~Capp\'e.
\newblock The {KL-UCB} algorithm for bounded stochastic bandits and beyond.
\newblock In \emph{Proc. of Conference On Learning Theory (COLT)}, 2011.

\bibitem[Graves and Lai(1997)]{graves1997}
T.~L. Graves and T.~L. Lai.
\newblock Asymptotically efficient adaptive choice of control laws in
  controlled markov chains.
\newblock \emph{SIAM J. Control and Optimization}, 35\penalty0 (3):\penalty0
  715--743, 1997.

\bibitem[Kleinberg(2004)]{Kleinberg2004}
R.~Kleinberg.
\newblock Nearly tight bounds for the continuum-armed bandit problem.
\newblock In \emph{Proc. of the conference on Neural Information Processing
  Systems (NIPS)}, 2004.

\bibitem[Kleinberg et~al.(2008)Kleinberg, Slivkins, and Upfal]{kleinberg2008}
R.~Kleinberg, A.~Slivkins, and E.~Upfal.
\newblock Multi-armed bandits in metric spaces.
\newblock In \emph{Proc. of the 40th annual ACM {S}ymposium on {T}heory of
  {C}omputing (STOC)}, pages 681--690, 2008.

\bibitem[Lai(1987)]{lai1987}
T.~L. Lai.
\newblock Adaptive treatment allocation and the multi-armed bandit problem.
\newblock \emph{The Annals of Statistics}, 15\penalty0 (3):\penalty0
  1091--1114, 09 1987.

\bibitem[Lai and Robbins(1985)]{lai1985}
T.L. Lai and H.~Robbins.
\newblock Asymptotically efficient adaptive allocation rules.
\newblock \emph{Advances in Applied Mathematics}, 6\penalty0 (1):\penalty0
  4--2, 1985.

\bibitem[M\"{u}ller and Stoyan(2002)]{MullerStoyan}
A.~M\"{u}ller and D.~Stoyan.
\newblock \emph{{Comparison Methods for Stochastic Models and Risks}}.
\newblock Wiley, 2002.

\bibitem[Slivkins(2011)]{slivkins11}
A.~Slivkins.
\newblock Contextual bandits with similarity information.
\newblock In \emph{Proc. of Conference On Learning Theory (COLT)}, pages
  679--702, 2011.

\bibitem[Wets(1985)]{Wets1985}
R.~Wets.
\newblock On the continuity of the value of a linear program and of related
  polyhedral-valued multifunctions.
\newblock \emph{Mathematical Programming Study}, 1985.

\end{thebibliography}

\appendix

\section{Proof of Theorem \ref{th:low}}

To establish the asymptotic lower bound, we apply the techniques used in \cite{graves1997} to investigate efficient adaptive decision rules in controlled Markov chains. We recall here their general framework. Consider a controlled Markov chain $(X_t)_{t\ge 0}$ on a finite state space ${\cal S}$ with a control set $U$. The transition probabilities given control $u\in U$ are parametrized by $\theta$ taking values in a compact metric space $\Theta$: the probability to move from state $x$ to state $y$ given the control $u$ and the parameter $\theta$ is $p(x,y;u,\theta)$. The parameter $\theta$ is not known. The decision maker is provided with a finite set of stationary control laws $G=\{g_1,\ldots,g_K\}$ where each control law $g_j$ is a mapping from ${\cal S}$ to $U$: when control law $g_j$ is applied in state $x$, the applied control is $u=g_j(x)$. It is assumed that if the decision maker always selects the same control law $g$, the Markov chain is irreducible with stationary distribution $\pi_\theta^g$. Now the expected reward obtained when applying control $u$ in state $x$ is denoted by $r(x,u)$, so that the expected reward achieved under control law $g$ is: $\mu_\theta(g)=\sum_xr(x,g(x))\pi_\theta^g(x)$. There is an optimal control law given $\theta$ whose expected reward is denoted $\mu_\theta^{\star}\in \arg\max_{g\in G} \mu_\theta(g)$. Now the objective of the decision maker is to sequentially select control laws so as to maximize the expected reward up to a given time horizon $T$. As for MAB problems, the performance of a decision scheme can be quantified through the notion of regret which compares the expected reward to that obtained by always applying the optimal control law.

We now apply the above framework to our Lipschitz bandit problem, and we consider $\theta \in {\Theta}_L$. The Markov chain has values in $\{0,1\}$. The set of control laws is $G=\{1,\ldots,K\}$. These laws are constant, in the sense that the control applied by control law $k$ does not depend on the state of the Markov chain, and corresponds to selecting arm $k$. The transition probabilities are: 
$$
p(x,y;k,\theta) = \left\{
\begin{array}{ll}
\theta_k, & \hbox{if }y=1,\\
1-\theta_k, & \hbox{if }y=0.
\end{array}
\right.
$$
Finally, the reward $r(x,k)$ is just given by the state $x$. 

We now fix $\theta \in {\Theta}_L$. Define the set $B(\theta)$ consisting of all {\it bad} parameters $\lambda \in {\Theta}_L$ such that $k^{\star}$ is not optimal under parameter $\lambda$, but which are statistically {\it indistinguishable} from $\theta$:
$$
B(\theta)=\{ \lambda \in {\Theta}_L : \lambda_{k^{\star}} = \theta_{k^{\star}} \textrm{ and }  \max_k  \lambda_k > \lambda_{k^{\star}}  \},
$$
$B(\theta)$ can be written as the union of sets $B_k(\theta)$, $k \in {\cal K}^-$ defined as:
$$
B_k(\theta)=\{ \lambda \in B(\theta) :  \lambda_k > \lambda_{k^{\star}}\}.
$$
By applying Theorem 1 in \cite{graves1997}, we know that $C(\theta)$ is the minimal value of the following LP:
\begin {eqnarray}
\textrm{min }  & \sum_k c_k(\theta^{\star}  - \theta_k) \\
\textrm{s.t. } & \inf_{\lambda \in B_k(\theta)} \sum_{l \in {\cal K}} c_l I(\theta_l,\lambda_l) \geq 1,\ \forall k\in {\cal K}^- \label{eq:con1}\\
& c_k \geq 0, \quad \forall k \in {\cal K}.
\end {eqnarray}

To conclude the proof, it is sufficient to remark that for any $k$,
$$
\inf_{\lambda \in B_k(\theta)} \sum_{l \in {\cal K}} c_l I(\theta_l,\lambda_l)=\sum_{l \in {\cal K}} c_l I(\theta_l,\lambda_l^k),
$$
which is easy in view of the definition of $\lambda^k$, by monotonicity of $x\mapsto I(\theta_k,x)$ when $x\ge \theta_k$.
\ep

\section{Proof of Theorem \ref{th:kl_concentr}}

In this section, we first establish the concentration inequality assuming that Lemma \ref{lem:peeling} holds. We then prove Lemma \ref{lem:peeling}, and to this aim, we state and use two further intermediate results, Lemmas \ref{lem:martingale} and \ref{lem:exp_moments}, proved at the end of this section. Without loss of generality, we assume that $t_k(n)\ge 1$ for any $k$ (the case where for some $k$, $t_k(n)=0$ is treated similarly).

\medskip
\noindent
{\it Proof of Theorem \ref{th:kl_concentr}.} Let $\delta\ge K+1$ and $\eta > 0$. Define $D = \ceil{ \log(n) / \log(1 + \eta) }$, and the set ${\cal D} = \{1,\dots,D\}^K$. Introduce the following events:
  \als{
  A &= \left\{ \sum_{k=1}^K  t_k(n) I^{+}(\hat\theta_k(n) , \theta_k   ) \geq \delta \right\}, \sk
  B_{d} &= \cap_{k=1}^K \left\{ (1+\eta)^{d_k-1} \leq t_k(n) \leq (1+\eta)^{d_k} \right\}, \quad \hbox{for all }d\in {\cal D}.
  }
We have $A=\cup_{d\in {\cal D}}(A\cap B_d)$, and hence $\PP[A] \leq \sum_{d \in {\cal D}}  \PP[ A \cap B_{d}]$. We let $\eta= 1/(\delta-1)$ and apply Lemma~\ref{lem:peeling} with $\overline{t}_k=(1+\eta)^{d_k-1}$. Since $\delta\ge K+1$, for $\eta= 1/(\delta-1)$, $\delta\ge (1+\eta)K$, and the application of Lemma \ref{lem:peeling} is legitimate. We obtain for all $d\in {\cal D}$: 
$$
\PP[ A \cap B_{d}] \leq \left(\frac{ \delta e }{K}\right)^K  e^{-\delta/(1+\eta)}.
$$
Since $|{\cal D}| = D^K$, we deduce that $\PP[A] \leq  \left( \frac{ D \delta e}{K}\right)^K  e^{-\delta/(1+\eta)}$. Now with our choice  $\eta = 1/(\delta-1)$, and using the inequality $\log(1+\eta) = - \log(1/(1+\eta)) \geq 1 - 1/(1+\eta) = 1/\delta$, we get:
	\eqs{
	\PP[A] \leq e^{-\delta} \left( \frac{\delta \ceil{\delta \log(n)}}{K}\right)^K e^{K+1},
	}
which concludes the proof.	
\ep

\begin{lemma}\label{lem:peeling}
For any $k=1,\ldots,K$, let $1 \leq \bar{t}_k \leq n$. Let $\eta > 0$. Define the event:
	\eqs{
	C = \cap_{k=1}^K \{ \bar{t}_k \leq t_k(n) \leq (1+\eta) \bar{t}_k  \}.
	}
For $\delta \geq (1+\eta)K$, we have:
	\eqs{
	\PP\left[  \indic_{C}  \sum_{k=1}^K t_k(n) I^+( \hat\theta_k(n),  \theta_k )   \geq \delta \right] \leq \left(\frac{ \delta e }{K}\right)^K  e^{-\delta/(1+\eta)}.
	}
\end{lemma}

\medskip
\noindent
{\it Proof of Lemma \ref{lem:peeling}.} 
Define the event $E = \{ \indic_{C}  \sum_{k=1}^K t_k(n) I^+( \hat\theta_k(n),  \theta_k )   \geq \delta \}$. We shall prove that for all $\zeta \in (\RR^+)^K$:
$$
\PP[ \cap_{k=1}^K \{  \indic_C t_k(n) I^{+}( \hat\theta_k(n) ,  \theta_k ) \geq \zeta_k  \} ] \leq e^{- (\sum_{k=1}^K \zeta_k)/(1+\eta) }.
$$
Let $\zeta \in (\RR^+)^K$. For $t \geq 0$, we define $x_{k}(t)$ such that (i) if there exists $0\le x\le \theta_k$ such that $t I^{+}( x ,\theta_k ) = \zeta_k$, then $x_k(t)=x$, (ii) else $x_k(t)=0$. By monotonicity of $I^{+}$, $t \mapsto x_{k}(t)$ is increasing. Hence $t_k(n) I^{+}( \hat\theta_k(n),\theta_k ) \geq \zeta_k$ implies that $\hat\theta_k(n) \leq x_{k}(t_k(n)) \leq  x_{k}(\bar{t}_k(1+\eta))$. We also have $\bar{t}_k I^{+}( x_{k}(\bar{t}_k(1+\eta)) ,\theta_k ) = \zeta_k/(1+\eta)$.
	
We deduce that	
\als{
\PP[ \cap_k  \{ \indic_C  t_k(n) I^{+}( \hat\theta_k(n) ,  \theta_k ) \geq \zeta_k  \} ] & \leq \PP[ \cap_k \{ \hat\theta_k(n) \leq x_{k}(t_k(n)) ,  C \} ]\sk
&\leq \PP[ \cap_k \{ \hat\theta_k(n) \leq x_{k}(\bar{t}_k(1+\eta)) ,  C \} ]\sk
&\leq \prod_{k=1}^K e^{- \bar{t}_k  I(x_{k}(\bar{t}_k(1+\eta)), \theta_k)} ] = e^{- \sum_{k=1}^K \zeta_k/(1+\eta) },
}	
where the last inequality is obtained by applying Lemma~\ref{lem:martingale} with $C_k=x_{k}(\bar{t}_k(1+\eta))$. Next we apply Lemma~\ref{lem:exp_moments} with $Z_k= \indic_C t_k(n) I^{+}( \hat\theta_k(n),\theta_k )$ and  $a=1/(1+\eta)$. We get:
	\als{
	\PP[ E ] &\leq \left(\frac{ \delta e }{K (1+\eta)}\right)^K  e^{-\delta/(1+\eta)}, \leq \left(\frac{ \delta e }{K}\right)^K  e^{-\delta/(1+\eta)}.  
	}
\ep

\begin{lemma}\label{lem:martingale}
For any $k=1,\ldots,K$, let $1 \leq \bar{t}_k \leq n$. Then for all $0\le C_k \le \theta_k$ we have:
	\eqs{
	\PP[ \cap_{k=1}^K \{  \hat\theta_k(n) \leq C_k , \bar{t}_k \leq t_k(n)  \}  ] \leq \prod_{k=1}^K e^{- \bar{t}_k  I( C_k, \theta_k)}.
	}
\end{lemma}

\medskip
\noindent
{\it Proof of Lemma \ref{lem:martingale}.} 
For all $k=1,\ldots,K$ and $\lambda$, we define 
\eqs{
\phi_k(\lambda) = \log(  \EE[ e^{\lambda X_k(1)} ] ) = \log( \theta_k e^{\lambda} + (1-\theta_k) )  .
}
One can easily show that for all $x \in [0, \theta_k]$, $I(x,\theta_k) = \sup_{\lambda \leq 0} \{ \lambda x - \phi_k(\lambda) \}$. Define the events $F = F_1 \cap F_2$, where $F_1 = \cap_{k=1}^K \{ \bar{t}_k \leq t_k(n) \}$, and  $F_2 = \cap_{k=1}^K \{  \hat\theta_k(n) \leq C_k \}$.

For all $k$, let $\lambda_k \leq 0$, and define $G(n) = \exp \left(  \sum_{k=1}^{K} \lambda_k S_k(n) - t_k(n) \phi_k(\lambda_k) \right)$. For all $n' \leq n$ we have $G(n') = G(n'-1)  \prod_{k=1}^K e^{ B_k(n') (\lambda_k X_k(n') - \phi_k(\lambda_k) )}$. Since $B_k(n')$ is ${\cal F}_{n' - 1}$ measurable and $\{ X_k(n')\}_k$ is independent of ${\cal F}_{n' - 1}$, we deduce that $\EE[G(n') | {\cal F}_{n' - 1}] = G(n'-1)$, i.e., $G$ is a martingale. Furthermore $\EE[G(n)] = 1$.

	For all $k$, we set
	\eq{
	\lambda_k = \arg \max_{\lambda \leq 0}  \{ \lambda C_k - \phi_k(\lambda) \},
	}
so that $\lambda_k C_k - \phi_k(\lambda_k) = I(C_k , \theta_k )$. We have $\lambda_k < 0$ and therefore:
	\als{
	\PP[F ] &= \PP[  \cap_{k=1}^{K} \{  S_k(n)   \leq  t_k(n) C_k \,,\, F_1 \} ] \sk
	& \leq \PP[\sum_{k=1}^{K} \lambda_k S_k(n)   \geq   \sum_{k=1}^{K} \lambda_k t_k(n) C_k \,,\, F_1 ] \sk
	&\leq \PP[ \indic_{F_1} e^{ \sum_{k=1}^{K} \lambda_k S_k(n) }  \geq   e^{\sum_{k=1}^{K} \lambda_k t_k(n) C_k}  ] \sk
	&=\PP[  \indic_{F_1} G(n)  \geq  e^{  \sum_{k=1}^{K} t_k(n) ( \lambda_k C_k - \phi_k(\lambda_k) )} ] \sk
	&=\PP[  \indic_{F_1} G(n)  \geq  e^{  \sum_{k=1}^{K} t_k(n) I(C_k,\theta_k)} ] \sk
	& \leq \PP[  \indic_{F_1} G(n)  \geq e^{  \sum_{k=1}^{K} \bar{t}_k I(C_k,\theta_k) } ].
	}
Using Markov inequality and the fact that $\EE[\indic_{F_1} G(n)] \leq  \EE[G(n)] = 1$, and we obtain the announced result:
	\als{
	\PP[ F ] &\leq  \EE[\indic_{F_1} G(n)] e^{ - \sum_{k=1}^{K} \bar{t}_k I(C_k,\theta_k) } \leq  e^{ -\sum_{k=1}^{K} \bar{t}_k I(C_k,\theta_k) }.
	}
\ep

\begin{lemma}\label{lem:exp_moments}
Let $a > 0$, $K \ge 2$. Let $Z \in \RR^K$ be a random variable such that for all ${\zeta} \in (\RR^+)^K$:
\eqs{
\PP[ Z \geq {\zeta} ] \leq e^{-a \sum_{k=1}^K \zeta_k}.
}
Then for all $\delta \geq K/a \in \RR^+$:
\eqs{
	\PP[ \sum_{k=1}^K Z_k \geq  \delta] \leq  \left(\frac{ a \delta e }{K}\right)^K  e^{-a \delta}.
}
\end{lemma}

\medskip
\noindent
{\it Proof of Lemma \ref{lem:exp_moments}.} 
Let $Y \in (\RR^+)^K$ a vector whose components are independent and exponentially distributed with parameter $a$. Then, $Z \leq_{uo} Y$ since for all $\zeta \in (\RR^+)^K$ (see Lemma \ref{lem:stochastic_comparison}):  
\eqs{
	\PP[  Z \geq \zeta] \leq e^{-a \sum_{k=1}^K \zeta_k} = \PP[Y \geq \zeta].
	}

Let $\lambda \in [0,a)$ and $\delta \in \RR^+$. Using Markov inequality we get:
	\als{
		\PP[ \sum_{k=1}^K Z_k \geq  \delta] & = \PP[ e^{ \lambda \sum_{k=1}^K Z_k} \geq  e^{\lambda \delta}] \leq e^{- \lambda  \delta} \EE[ e^{ \lambda \sum_{k=1}^K Z_k} ] \sk
		&= e^{- \lambda \delta} \EE[ \prod_{k=1}^K e^{ \lambda  Z_k} ] \leq e^{- \lambda \delta} \EE[ \prod_{k=1}^K e^{ \lambda  Y_k} ] \sk
		&= e^{- \lambda \delta} \prod_{k=1}^K \EE[  e^{ \lambda  Y_k} ] .
	}
where we have used the results of Lemma~\ref{lem:stochastic_comparison} with $f_k(z)=e^{z \lambda}$ for all $k$. Note that $z \mapsto e^{z \lambda}$ is positive and increasing.
	
Furthermore we have $\EE[ e^{ \lambda Y_k} ] = \int_{0}^{+\infty} a e^{-ay}  e^{\lambda y}  dy = \frac{a}{a - \lambda}$. 
Hence we have established that for all $0\le \lambda < a$:
	\eqs{
	\PP[ \sum_{k=1}^K Z_k \geq  \delta] \leq e^{- \lambda \delta} \frac{a^K}{(a - \lambda)^K}.
	} 
Setting $\lambda = a - K/\delta \geq 0$, we obtain:
	\eqs{
	\PP[ \sum_{k=1}^K Z_k \geq \delta] \leq  \left(\frac{ a \delta e }{K}\right)^K  e^{-a \delta}.
	}
\ep

The next lemma presents a result on multivariate stochastic ordering, see \cite{MullerStoyan}[Theorem 3.3.16].
\begin{lemma}\label{lem:stochastic_comparison}
Let $X$ and $Y$ be two random variables on $\RR^K$. The following are equivalent:

(i) $X \leq_{uo} Y$,

(ii) For all $x \in \RR^K$, $\PP[ X \geq x] \leq \PP[ Y \geq x]$,

(iii) For all collections of non negative increasing functions $f_1,\dots,f_K$ we have $\EE[\prod_{k=1}^K f_k(X_k)] \leq \EE[\prod_{k=1}^K f_k(Y_k)]$.
\end{lemma}

\section{Proof of Theorem \ref{th:oslbperf}}

We first present two important corollaries of our concentration inequality (Theorem \ref{th:kl_concentr}). 

\begin{corollary}\label{coro1}
Let $f(n)=\log(n)+(3K+1)\log\log(n)$. There exists $n_0$ such that for all $n\ge n_0$:
$$
\mathbb{P}\left[ \sum_{k=1}^K t_k(n)I^+(\hat\theta_k(n),\theta_k)\ge f(n)\right] \le {1\over n\log(n)}.
$$
\end{corollary}

\begin{corollary}\label{coro2}
Let $f(n) = \log(n) + (3K+1) \log\log(n)$, and define $\lambda_{k'}^{q,k} = q - L|x_k-x_{k'}|$. Then there exists $n_0$ such that for all $n \geq n_0$:
\eqs{
\PP[ b_k(n) < \theta_k ] \leq {1\over n\log(n)}.
}	
\end{corollary}

\medskip
\noindent
{\it Proof of Corollary \ref{coro2}.}
Since $I^+$ is increasing in its second argument, the event $b_k(n) < \theta_k$ implies that:
 \eqs{
 \sum_{k'=1}^{K} t_{k'}(n) I^{+}(\hat\theta_{k'}(n) ,  \lambda_{k'}^{\theta_k , k}) \geq f(n).
 }
 Furthermore, by definition $\lambda_{k'}^{\theta_k , k} = \theta_k - L|x_k-x_{k'}| \leq \theta_{k'} $. Hence:
 \eqs{
 \sum_{k'=1}^{K} t_{k'}(n) I^{+}(\hat\theta_{k'}(n) ,  \theta_{k'}) \geq f(n).
 }
 We can now apply Corollary~\ref{coro1} and obtain:
 \als{
 \PP[b_k(n) < \theta_k ] &\leq \PP[ \sum_{k'=1}^{K} t_{k'}(n) I^{+}(\hat\theta_{k'}(n) ,  \theta_{k'}) \geq f(n)  ] \leq {1\over n\log(n)}.
 }
\ep

We then give an important lemma that allows us to upper bound the average cardinalities of particular sets of rounds. This lemma is stated and proved in \cite{techreport}.
\begin{lemma}\label{lem:deviation}
Let $k\in {\cal K}$, and $\epsilon > 0$. Define ${\cal F}_n$ the $\sigma$-algebra generated by \\
$( X_k(t) )_{1 \leq t \leq n, 1 \leq k \leq K}$. Let $\Lambda \subset \NN$ be a (random) set of instants. Assume that there exists a sequence of (random) sets $(\Lambda(s))_{s\ge 1}$ such that (i) $\Lambda \subset \cup_{s \geq 1} \Lambda(s)$, (ii) for all $s\ge 1$ and all $n\in \Lambda(s)$, $t_k(n) \ge \epsilon s$, (iii) $|\Lambda(s)| \leq 1$, and (iv) the event $n \in \Lambda(s)$ is ${\cal F}_n$-measurable. Then for all $\delta > 0$:
\eq{\label{eq:ineq1}
\EE[ \sum_{n \geq 1} \indic\{ n \in \Lambda , | \hat\theta_k(n) - \theta_k | > \delta \} ]  \leq  \frac{1}{\epsilon \delta^2}.
}
\end{lemma}

We are now ready to analyze the regret achieved under OSLB($\epsilon$).

\medskip
\noindent
{\it Proof of Theorem \ref{th:oslbperf}.} 
Let $S(\theta)$ denote the set of solutions of (\ref{eq:opt1}) for a given $\theta$. For any $\chi >0$, we define the set 
$$\Gamma_{\chi,\theta}=\cup_{ \{ \theta': |\theta_k'-\theta_k|<\chi, \forall k\} } S(\theta'),
$$
and for all $k$, $c_k^\chi=\sup\{ c_k: c\in \Gamma_{\chi,\theta}\}$. In view of Lemma~\ref{lem:LPcont}, $\theta' \mapsto  S(\theta')$ is upper hemicontinuous at $\theta$ and by Assumption~\ref{assum:uniqueLP} $S(\theta)$ reduces to a point. Therefore, for any open neigbourhood ${\cal B}$ of $S(\theta)$, there exists $\chi>0$ such that $S(\theta') \subset {\cal B}$ if $\sup_k |\theta'_{k} - \theta_k|< \chi$. Hence for all $k$: $c_k^{\chi} \to c_k(\theta)$, as $\chi \to 0$.

Fix $0 < \delta < (\theta^\star - \max_{k \neq k^\star} \theta_k)/2 $ and $\epsilon > 0$. To simplify the notation, we replace $\epsilon$ by $K\epsilon$ in the Theorem \ref{th:oslbperf}, and prove the result for this choice of $\epsilon$. 

Let $k$ be a suboptimal arm. We derive an upper on the number of times it is played. Let $n$ be a round where $k$ is played, i.e., $k(n)=k$. In view of the design of OSLB($\epsilon$), there are three possible scenarios: (a) $k$ can be the leader and its empirical reward exceeds the indexes of other arms, $L(n)=k$ and $\hat\theta_k(n)\ge \max_{l} b_l(n)$; (b) $k$ and $k^\star$ are not the leader, and $k$ can be either $\underline{k}(n)$ or $\overline{k}(n)$; (c) $k^\star$ is the leader, and again $k$ can be either $\underline{k}(n)$ or $\overline{k}(n)$. We investigate all cases, but we start by defining sets of rounds whose average cardinalities can be easily controlled:     
\als{ 
 A_k &= \{ 1 \leq n \leq T : k(n)=k, b_k(n) \leq \theta_k \} \sk
 B_k&= \{  n  \geq 1: k(n)=k , \min_{k'} t_{k'}(n) \geq \epsilon t_k(n), \max_{k'} | \hat\theta_{k'}(n) - \theta_{k'}  | \geq \delta    \}\sk 
 E_k&= \{  n  \geq 1: k(n)=k , | \hat\theta_{k}(n) - \theta_{k}  | \geq \delta    \}\sk
 F_k&= \{  n  \geq 1: k(n)=k , t_k(n)\le \min(t_{k'}(n),t_{k^\star}(n)), \max_{l\in\{k',k^\star\} } | \hat\theta_{l}(n) - \theta_{l}  | \geq \delta  \}
 }
and $A = \cup_{k} A_k$, $B = \cup_{k} B_k$, $E = \cup_{k} E_k$, $F = \cup_{k} F_k$. From the concentration inequality, and its corollaries, we have $\EE[|A|] \leq C_1 \log\log(T)$. We use Lemma \ref{lem:deviation} to bound the cardinalities of the other sets. 
\begin{itemize}
\item Bound for $B_k$. Let us fix $k'\neq k$. We apply Lemma \ref{lem:deviation} to $k'$ with $\Lambda(s)=\{ n: k(n)=k, \min_l t_l(n)\geq \epsilon s, t_k(n)=s\}$, and $\Lambda = \cup_s\Lambda(s)$. We get that:
$$
\mathbb{E}\left[ \big |\{n:k(n)=k,  \min_l t_l(n)\geq \epsilon t_k(n), | \hat\theta_{k'}(n) - \theta_{k'}  | \geq \delta \} \big| \right] \le {1\over \epsilon\delta^2}.
$$ 
We conclude that: $\mathbb{E}[| B_k | ]\le K/(\epsilon\delta^2)$.
\item Bound for $E_k$. The application of lemma is direct here, and we get: $\mathbb{E}[| E_k| ]\le 1/\delta^2$.
\item Bound for $F_k$. Using the same argument as that used to bound the cardinality of $B_k$, we get: $\mathbb{E}[ |F_k |]\le 2/\delta^2$.
\end{itemize}

Next we consider $n \notin A \cup B\cup E\cup F$ such that $k$ is played. We treat all cases (a), (b), and (c) that can arise in such a round.\\

\medskip
\noindent
\underline{Case (a)} We assume here that $k=L(n)$ and that $k(n)=k$, so that $\hat\theta_k(n)\ge \max_l b_l(n)$. Hence, since $n\notin A_{k^\star}$, $\hat\theta_k(n)\ge b_{k^\star}(n)\ge \theta^\star$. In summary, $\hat\theta_k(n)\ge \theta^\star$, which is impossible because of our choice of $\delta$  ($< \theta^\star - \theta_k$), and $n\notin E_k$.

\medskip
\noindent
\underline{Case (b)} Let $k'\notin \{k,k^\star\}$ be the leader in round $n$, and assume that $k(n)=k$. We consider two subcases: (i) $k=\underline{k}(n)$, and (ii) $k=\overline{k}(n)$.\\
(i) In this case, $k$ has been played less than any other arm, and so $t_k(n)\le \min(t_{k'}(n),t_{k^\star}(n))$. On the other hand, since $k'$ is the leader, we have $\hat\theta_{k'}(n)\ge \hat\theta_{k^\star}(n)$, which implies that either $\theta_{k'}$ or $\theta_{k^\star}$ is badly estimated. More precisely, we proved that $n\in F_k$, which is impossible.\\
(ii) In this case, we know that $t_k(n)\le t_{\underline{k}(n)}(n)/\epsilon$. In addition, again, we have $\hat\theta_{k'}(n)\ge \hat\theta_{k^\star}(n)$, and so either $\theta_{k'}$ or $\theta_{k^\star}$ is badly estimated. We proved that $n\in B_k$, which is impossible. 

\medskip
\noindent
\underline{Case (c)} Assume that $k^\star = L(n)$. $k$ is played, and we need to consider two subcases: (i) $k=\underline{k}(n)$, and (ii) $k=\overline{k}(n)$.\\  
(i) In this case, since $k=\underline{k}(n)$, we have $t_k(n)\le \min_l t_l(n)$, and hence $\epsilon t_k(n)\le \min_l t_l(n)$. Since $n\notin B_k$, in view of the previous inequality, all arms must be well-estimated, i.e., $\max_{l} | \hat\theta_{l}(n) - \theta_{l}  | < \delta$. This implies that for all $l\in {\cal K}$, $\hat{c}_l(n) \le c_l^\delta$. Now by definition in our algorithm, if $k(n)=k=\underline{k}(n)$, then $t_k(n) < \epsilon t_{\overline{k}(n)}(n)$, and so $t_k(n) <\epsilon \max_{l} c_l^\delta \log(n)$. In other words, $n\in D_k$ where 
$$
D_k = \{ 1 \leq n \leq T, n \notin A \cup B\cup E\cup F, L(n) = k^\star, k(n)=k,  t_k(n) \leq \epsilon \max_{k'} c_{k'}^\delta \log(T) \}.
$$
We shall bound the size of $D_k$ later in the proof.\\
(ii) In this case, we must have $t_{\underline{t}(n)}(n)\ge \epsilon t_k(n)$. Hence since $n\notin B_k$, all arms are well estimated, and hence again, for all $l\in {\cal K}$, $\hat{c}_l(n) \le c_l^\delta$. In particular, since $k$ is played, $t_k(n)\le c_k^\delta \log(n)$, and thus $n\in C_k$ where
$$
C_k = \{ 1 \leq n \leq T, n \notin A \cup B,  k(n) = k,  t_k(n) \leq c_k^\delta \log(T) \}.
$$

Nest we bound the expected cardinalities of $C_k$ and $D_k$. Since $t_k(n)$ is incremented if $n \in C_k$ or $n \in D_k$, we simply have:
$$
|C_k| \leq c_k^\delta \log(T), \quad |D_k| \leq \epsilon \max_{k'} c_{k'}^\delta \log(T).
$$

Putting it all together we have proven the announced regret bound:
\als{
R^{\pi}(T) &\leq \sum_{k \neq k^\star} (\theta^\star - \theta_k) (\EE[|C_k|] + \EE[|D_k|]) \sk
			& + \EE[|A|] + \EE[|B|] + \EE[|E|] + \EE[|F|] , \sk
			& \leq \log(T) \sum_{k \neq k^\star} (\theta^\star - \theta_k) ( c_k^\delta + \epsilon \max_{k'} c_{k'}^\delta) \sk 
			& + C_1 \log\log(T) + K^2 \epsilon^{-1}\delta^{-2} + 3K \delta^{-2}.
}
This completes the proof (because of our particular choice of $\epsilon$, and $\max_{l} c_{l}^\delta \le \sum_{l}c_l^\delta$).
\ep

\subsection{Continuity of solutions to parametric linear programs}
We state and prove Lemma~\ref{lem:LPcont}, a technical result about the continuity of the solutions of a parametric linear program with respect to its parameters. It follows from the general conditions of~\cite{Wets1985}.
\begin{lemma}\label{lem:LPcont}
Consider $A \in (\RR^+)^{K \times K}$ , $c \in (\RR^+)^K$, and ${\cal T} \subset (\RR^+)^{K \times K} \times (\RR^+)^{K}$. Define $t = (A,c)$. Consider the function $Q$ and the set-valued map $Q^\star$ \als{
Q(t) &= \inf_{x \in \RR^K} \{ c x | A x \geq 1 , x \geq 0 \} \sk
Q^\star(t) &= \{x: c x  \leq Q(t)   | A x \geq 1 , x \geq 0 \}.
}
Assume that:
\begin{itemize}
\item[(i)] For all $t \in {\cal T}$, all rows and columns of $A$ are non-identically $0$
\item[(ii)] $\min_{t \in {\cal T}} \min_k c_k > 0$
\end{itemize}
Then:
\begin{itemize}
\item[(a)] $Q$ is continuous on ${\cal T}$.
\item[(b)] $Q^\star$ is upper hemicontinuous on ${\cal T}$. 
\end{itemize}
\end{lemma}
\bp
Define \eqs{
c_0 =  \min(1, \min_{t \in {\cal T}} \min_k c_k) > 0,
}
and $a = \max_{(k,k')} A_{k,k'}$. Define the sets ${\cal K} = \{  x | A x \leq 1 \}$, ${\cal D} = \{  x | A x \leq c \}$ and ${\cal B} = [0,c_0/(aK)]^K$.
Then ${\cal B} \subset {\cal K} \cap {\cal D}$, so that both ${\cal K}$ and ${\cal D}$ have non-empty interior.  
By \cite{Wets1985}[Corollary 7], $t \to {\cal K}$ and $t \to {\cal D}$ are continuous on ${\cal T}$ since they have non-empty interior and all rows of $(A,1)$ and columns of $\binom{A}{c}$ are non identically $0$.
By \cite{Wets1985}[Theorem 2], $Q$ is continuous on ${\cal T}$ since both ${\cal K}$ and ${\cal D}$ are continuous on ${\cal T}$, proving the first statement.

Consider a sequence $\{ (t^i,x^i) \}_{i \geq 1}$, such that $x^i \in Q^\star(t^i)$ and $(t^i , x^i) \to (t,\overline{x})$, $i \to \infty$. Since for all $i \geq 1$ $c x^i \leq Q(t^i)$ and $A x^i \geq 1$ we have, by continuity, $A \overline{x} \geq 1$ and $c \overline{x} = Q(t)$ and so $\overline{x} \in Q^\star(t)$. Hence $Q^\star$ is upper hemicontinuous.
\ep

\section{Proof of Theorem~\ref{th:rrucbperf}}

\subsection{Proof of (i)}

Let $0 < \delta < (\theta^* - \max_{k \in {\cal K}^-} \theta_k)/2$ fixed throughout the proof. 
Define the random sets of rounds: $B = \{ 1 \leq n  \leq T: b_{k^\star}(n) \leq \theta^\star	\}$  the set of rounds at which the index of the optimal arm underestimates its true value $\theta^\star$, and $D_k = \{ n : k(n) = k , b_k(n) \geq \theta^\star - \delta 	\}$ the set of rounds at which $k$ is selected and its index is larger than $\theta^\star - \delta$.

Let $k \neq k^\star$ be a suboptimal arm, and let $n \notin B$ such that $k$ is selected $k(n)=k$. The possible events are:
\begin{itemize}
\item[(a)] If $L(n) \in \{ k,k^\star \}$ then $b_k(n) \geq b_{k^\star}(n) \geq \theta^\star$ since $n \notin B$, so $n \in D_k$.
\item[(b)] If $L(n) = k' \notin \{ k,k^\star \}$, then $b_k(n) \geq b_{k'}(n)$ and:
	\begin{itemize}	
		\item[(b-i)] If we further have $b_{k'}(n) \geq \theta^\star - \delta$ then $b_k(n) \geq \theta^\star - \delta$ so $n \in D_k$ as well.
		\item[(b-ii)] Otherwise $b_{k'}(n) \leq \theta^\star - \delta$.
	\end{itemize}
\end{itemize}
Define the random set of instants $E_{k} = \{ n \not\in B : k(n) = k, L(n) \not\in \{k,k^\star\}, b_{k^*}(n) > b_{L(n)}(n) , |\hat\theta_{k^*}(n) - \theta_{k^*} | \geq \delta    \}$. In the case (b-ii), we have $b_{L(n)}(n) \leq \theta^\star - \delta < \theta^\star \leq b_{k^\star}(n)$ since $n \notin B$. Also by definition of $L(n)$ we have that $\hat\theta_{k^\star}(n) \leq \hat\theta_{L(n)}(n) \leq b_{L(n)}(n) \leq \theta^\star - \delta$. So in case (b-ii) we have $n \in E_{k}$.

  In summary, $k(n)=k$ implies that $n \in B \cup E_{k} \cup D_k$ so:  $\EE[ t_k(T) ] \leq \EE[ |B| ] + \EE[ |E_{k}| ] + \EE[|D_k|]$. Let us upper bound the expected sizes of sets $B$, $E_{k}$ and $D_k$. 
  
\medskip
\noindent  
\underline{Expected size of $B$: } From Theorem~\ref{th:kl_concentr}, there exists a constant $C_1 \geq 0$ such that $\EE[|B|]$ is upper bounded by the Bertrand series:
 \eqs{
 \EE[|B|] \leq \sum_{n=1}^T C_1 (n\log(n))^{-1} \leq C_1 \log(\log(T)),
 }

\medskip
\noindent  
\underline{Expected size of $E_{k}$: } If $n \in E_k$ , we have $b_{k^*}(n) > b_{L(n)}(n) > \hat\theta_{L(n)}(n)$, so that by design of CKL-UCB, $k(n) \in \arg\min\limits_{k : b_{k}(n)  >  b_{L(n)}(n) } t_k(n)$ and  $k^* \in \{ k : b_{k}(n)  >  b_{L(n)}(n) \}$. Since $k(n)=k$, we have $t_k(n) \geq t_{k^\star}(n)$. Define $s = \sum_{n'=1}^n 1 \{  n' \in E_k\}$, this implies $t_{k^*}(n) \geq s$. Applying Lemma \ref{lem:deviation} as earlier, we conclude that $\EE[ E_k ] \leq \delta^{-2}$.

\medskip
\noindent  
\underline{Expected size of $D_k$: } Define $F_k^\delta =\{n : k(n) = k,\ |\hat\theta_k(n) -\theta_k|<\delta\}$ and $\overline{F_k^\delta} =\{n : k(n) = k,\ |\hat\theta_k(n) -\theta_k|\geq\delta\}$. Let us consider a round $n\in D_k\cap F_k^\delta$. Assume that $t_k(n) > f(n)/I(\theta_k+\delta, \theta^*-\delta)$. Since $n\in D_k$ and $k(n) = k$, we have: $b_k(n) \geq \theta^*-\delta$. Therefore, from the monotonicity of $I(x,y)$ in $y$ when $y>x$, we have:
\begin{equation}\label{eqContradiction}
t_k(n) I(\hat\theta_k(n), \theta^*-\delta) \leq \sum\limits_{i\in\mathcal{K}} t_i(n) I(\hat\theta_i(n), \lambda_i^{\theta^*-\delta, k} ) \leq \sum\limits_{i\in\mathcal{K}} t_i(n) I(\hat\theta_i(n), \lambda_i^{b_k(n), k}) = f(n)
\end{equation}
where the last equality comes from our definition of $b_k(n)$. Furthermore, by our assumption and since $\hat\theta_k(n)\leq \theta_k+\delta$ (since $n\in F_k^\delta$):
\eqs{
f(n) < t_k(n)I(\theta_k+\delta, \theta^*-\delta) \leq t_k(n) I(\hat\theta_k(n), \theta^*-\delta),
}
which contradicts \eqref{eqContradiction}.
Thus for all rounds in $n\in D_k\cap F_k^\delta$ we have $t_k(n) \leq f(n)/I(\theta_k+\delta, \theta^*-\delta)$ and consequently $\EE[|D_k|] \leq f(T)/I(\theta_k+\delta, \theta^*-\delta) + \EE[|\overline{F_k^{\delta}}|]$.

Again a direct application of Lemma \ref{lem:deviation} yields $\EE[|\overline{F_k^{\delta}}|] \leq \delta^{-2}$. Thus, we have:
\eqs{
\EE[ t_k(T) ] \leq f(T)/I(\theta_k+\delta, \theta^*-\delta) + C_1 \log(\log(T)) + 2 \delta^{-2}.
}

\ep

\subsection{Proof of (ii)}

We work with a fixed sample path throughout the proof. Since for all $k$ when $T\to\infty$ we have $t_k(T) \to \infty$ a.s., so by the law of large numbers $\hat\theta_k(T) \to\theta_k$ as ${T \to \infty}$. 

From the first statement of the theorem, we have that for all $k \neq k^\star$, $\lim \sup\limits_{T \to \infty} \EE[ t_k(T)] / \log(T )< \infty$  which implies that $\lim \sup\limits_{T \to \infty} t_k(T) / \log(T) < \infty$ . In turn we have that $t_{k^\star}(T) = T - \sum_{k \neq k^\star} t_k(T) = T - O(\log(T))$, so that $t_{k^\star}(T)/T \to_{T \to \infty} 1$. By Pinsker's inequality: 
\eqs{
\hat\theta_{k^\star}(T) \leq b_{k^\star}(T) \leq \hat\theta_{k^\star}(T) + \sqrt{2 f(T)  / t_{k^\star}(T) } 
}
and we can deduce $b_{k^\star}(T) \to \theta^\star$ as $T\to\infty$, because $\hat\theta_{k^\star}(T) \to \theta^\star$ and $f(T) / t_{k^\star}(T) = f(T)/( T - O(\log(T)) ) \to 0$ when $T\to\infty$.

Let $\delta$ such that $0 < \delta < (\theta^\star - \max\limits_{k \in {\cal K}^-} \theta_k)/2$, by the above reasoning there exists $n_0 \in \NN$ (depending on the sample path and $\delta$) such that for all $n \geq n_0$ we have  $|b_{k^\star}(n) - \theta^\star| \leq \delta$ and  $|\hat\theta_k(n) - \theta_k| \leq \delta$ for all $k$. It is noted that for all $n \geq n_0$, $L(n) = k^\star$, since $\delta < (\theta^\star - \max\limits_{k \neq k^\star} \theta_k)/2$.

Let $\alpha_0 \geq 0$, and assume that there exists $T$ large enough such that $t_k(T) = \alpha_0 f(T)$ and $\alpha_0 f(T) > t_k(n_0)$. Therefore there exists $n_0 \leq n \leq T$ such that $t_k(n) = \alpha_0 f(T) - 1$ and $k$ is selected at time $n$: $k(n) = k$. Define ${\cal N} = \{k': b_{k'}(n) \leq b_{k^\star}(n) \}$. Consider $k' \in {\cal N}$, since $n \geq n_0$, we have that $L(n) = k^\star$ and $b_{L(n)} \leq  \theta^\star + \delta$. So $b_{k'}(n) \leq \theta^\star + \delta$ which implies (by definition of $b_{k'}(n)$):
\eq{\label{eq:lp1}
	\sum_{k''\in  \mathcal{K}} t_{k''}(n) I(\theta_{k''}-\delta , \lambda_{k''}^{ \theta^\star + \delta,k'} ) \geq f(n)
}
Also, since $k(n) = k$, by design of CKL-UCB we have $t_k(n) = \arg\min\limits_{k' \not\in {\cal N} } t_{k'}(n)$, so that : 
\eq{ \label{eq:lp2}
t_{k'}(n) \geq t_k(n) = \alpha_0 f(T) - 1 \geq \alpha_0 f(n) - 1 , \forall k' \not\in {\cal N}.
}
Finally, since $k(n) = k$, $L(n) = k^\star$,  we must have $b_k(n) \geq b_{k^\star}(n) \geq \theta^\star - \delta$, so that:
\al{
	\sum_{k'\in \mathcal{K}} t_{k'}(n) I(\theta_{k'} + \delta , \lambda_{k'}^{ \theta^\star - \delta,k} ) &\leq f(n)  \sk
	(\alpha_0 f(T) - 1) I(\theta_{k} + \delta , \lambda_{k}^{ \theta^\star - \delta,k} ) + \sum_{k'\in \mathcal{K} \setminus\{k\}} t_{k'}(n) I(\theta_{k'} + \delta , \lambda_{k'}^{ \theta^\star - \delta,k} ) &\leq f(n) \sk
	(\alpha_0 f(n) - 1) I(\theta_{k} + \delta , \lambda_{k}^{ \theta^\star - \delta,k} ) + \sum_{k'\in \mathcal{K} \setminus\{k\}} t_{k'}(n) I(\theta_{k'} + \delta , \lambda_{k'}^{ \theta^\star - \delta,k} ) &\leq f(n) \label{eq:lp3}
}
Define the matrix $\tilde{A} = ( \tilde{a}_{k'k})_{k,k'}$, with  $\tilde{a}_{k'k} = I(\theta_{k'}+\delta , \lambda_{k'}^{ \theta^\star - \delta,k} )$ for all $k'$ and $\tilde{a}_{k''k'} =  I(\theta_{k''}-\delta , \lambda_{k''}^{ \theta^\star + \delta,k'} ) $ for all $k' \neq k$ and all $k''$. 

Define $\alpha_{k'}(n) = t_{k'}(n)/ f(n)$ for all $k'$, and by dividing equations \eqref{eq:lp3} , \eqref{eq:lp2} and \eqref{eq:lp1} by $f(n)$, we obtain:
\als{
 	\alpha_0 \tilde{a}_{kk} +  \sum_{k'\in \mathcal{K} \setminus\{k\}} \alpha_{k'}(n) \tilde{a}_{k'k} \leq 1, \hspace{2.5cm}\sk
	\alpha_{k'}(n) \geq \alpha_0-\frac{1}{f(n)},	 \hspace{1cm} \forall k' \notin {\cal N}, \sk
	\alpha_0 \tilde{a}_{kk'} + \sum_{k''\in \mathcal{K} \setminus\{k\}} \alpha_{k''}(n) \tilde{a}_{k''k'}  \geq 1  , \hspace{1cm} \forall k' \in {\cal N}.
}
It is noted that $\alpha_k' (n)\geq 0 $ for all $k'$ by definition. Define $\alpha = (\alpha_1,\dots,\alpha_K)$ a limit point of the sequence $( \alpha(n) )_{n \geq 1}$ (note that this sequence need not converge and might have several limit points). First letting $n \to \infty$ along a converging subsequence and then letting $\delta \to 0$ the constraints above become: 
\als{
 	\alpha_0 a_{kk} +  \sum_{k'\in \mathcal{K} \setminus\{k\}} \alpha_{k'} a_{k'k} \leq 1, \hspace{2.5cm}\sk
	\alpha_{k'} \geq \alpha_0,	 \hspace{1cm} \forall k' \notin {\cal N}, \sk
	\alpha_0 a_{kk'} + \sum_{k''\in \mathcal{K} \setminus\{k\}} \alpha_{k''} a_{k''k'}  \geq 1  , \hspace{1cm} \forall k' \in {\cal N}.
}
Therefore, by definition of $d_k$, we must have:
\als{
 	\alpha_0  \tilde{a}_{kk}  +  d_k( A,\alpha_0, {\cal N}) \leq 1,
 	}
and taking the infimum over ${\cal N}$ so that we obtain the condition:
\begin{equation}\label{eq:neccond}
	\alpha_0  \tilde{a}_{kk}  +  e_k(A,\alpha_0) \leq 1.
\end{equation}
Now consider $\alpha_0$ such that $\alpha_0  \tilde{a}_{kk}  +  e_k(A,\alpha_0) > 1$. Then in view of the necessary condition (\ref{eq:neccond}), we cannot have $t_k(T) \geq \alpha_0 \log(T)$, so that:
\eqs{
	\lim\sup_{T \to \infty} \frac{t_k(T)}{\log(T)} \leq \inf \{ \alpha_0 \geq 0:  a_{kk} \alpha_0 + e_k(A,\alpha_0)  > 1    \} = \beta_k(\theta). 
}
We get (ii) by Lebesgue's dominated convergence theorem, since $\sup_{T \geq 1} \EE[ \frac{t_k(T)}{\log(T)}] < \infty$ from $(i)$.

\subsection{Proof of (iii)}

In order to prove the last part of the theorem, it is sufficient to prove that for $\alpha_0 = 1/a_{kk}$, we have $e_k(A,\alpha_0 ) > 0$ so that $a_{kk} \alpha_0 + e_k(A,\alpha_0) = 1 + e_k(A,\alpha_0) > 1$ so that $\beta_k(\theta) < \alpha_0 = 1/a_{kk}$.

	We proceed by contradiction. Assume that $e_k(A,\alpha_0 ) = 0$. Then there exists ${\cal N}$ a subset of $\{ 1,\dots,K \} \setminus \{k^\star,k\}$ such that $d_k(A,\alpha_0, {\cal N}) = 0$. As a consequence there exists $\alpha_1,...,\alpha_{K}$ such that:
	
\als{  \sum_{k' \in \mathcal{K} \setminus\{k\} } & \alpha_{k'} a_{k'k} = 0 \sk
	\text{ s.t.\ }       &      \alpha_{k'} \geq \alpha_0,\;\;\; \forall k' \not\in {\cal N} \sk
	 										&		 \alpha_{k'} \geq 0,\;\;\; \forall k' \sk
	 									&			 \sum_{k''\in \mathcal{K} \setminus\{k\} } \alpha_{k''} a_{k''k'} \geq  1 - \alpha_0 a_{kk'},\;\;\; \forall k' \in {\cal N}.
}
Consider there exists $k'$ such that $a_{k'k} > 0$. Then we must have $k' \in {\cal N}$, otherwise $\alpha_{k'} \geq \alpha_{0} = 1/a_{kk}$ and $0 =  \sum\limits_{k'' \in \mathcal{K} \setminus\{k\} }\alpha_{k''} a_{k''k} \geq \alpha_{0} a_{k'k} > 0$, a contradiction. By the same reasoning we must also have $\alpha_{k'} = 0$.   

As said in the theorem statement, assume that there exists $k'\in \mathcal{N}$ such that $0 < a_{kk'} < a_{kk}$ and assume that for all $k''$ we have that if $a_{k''k} = 0$ then $a_{k''k'}$ as well. Considering $k'' = k'$ in our assumption, since $a_{k'k} = 0$ would imply $a_{k'k'} = 0$, we then have that $a_{k'k}>0$ since $a_{k'k'} = I(\theta_{k'}, \theta^*)>0$. From our previous argument we have that if $a_{k'k}>0$ then $k' \in \mathcal{N}$. Therefore :

\al{
\sum_{k'' \in \mathcal{K} \setminus\{k\}} \alpha_{k''} a_{k''k'} \geq  1 - \alpha_0 a_{kk'} = 1 -  a_{kk'} / a_{kk}  > 0\sk
\sum_{k'' \neq k,\ a_{k''k}  = 0} \alpha_{k''} a_{k''k'} > 0 \label{eq:constr}
}
By assumption, $a_{k''k}  = 0$ implies $a_{k''k'} = 0 $  so that the l.h.s. of~\eqref{eq:constr} is zero and cannot be strictly larger than 0. This is a contradiction, proving that $e_k(A,\alpha_0 ) = 0$ cannot occur and concludes the proof.

\ep

\end{document}